\pdfoutput=1
\documentclass[journal, 10pt]{IEEEtran}

\usepackage{amsmath}
\usepackage{fixmath}
\usepackage{amsfonts}

\usepackage{enumerate,letltxmacro}
\LetLtxMacro\itemold\item
\usepackage{amssymb}
\usepackage{setspace}

\usepackage{float}
\usepackage{subfig}
\usepackage{pbox}

\usepackage[table, rgb]{xcolor}
\usepackage{tikz, pgfplots}
\usepackage{verbatim}
\usepackage{cite,url,bm}
\usepackage{tablefootnote, threeparttable, booktabs, multirow}
\usepackage{algorithmicx}
\usepackage[ruled]{algorithm}
\usepackage{algpseudocode}

\alglanguage{pseudocode}
\usepackage{psfig}
\usepackage{latexsym}

\usepackage{graphicx}
\usepackage{tikz}

\usetikzlibrary{arrows,decorations.pathmorphing,backgrounds,positioning,fit,petri}

\usetikzlibrary{shapes,arrows}

 \pdfoutput=1
\usepackage[english]{babel}
\pgfplotsset{footnotesize}
\usepackage{parskip}
\usepackage{pgf}
\usepackage[normalem]{ulem}
\usepackage[inline]{enumitem}
\usepackage[autoplay]{animate}

\usepackage{lipsum}

\let\OLDthebibliography\thebibliography
\renewcommand\thebibliography[1]{
  \OLDthebibliography{#1}
  \setlength{\parskip}{0pt}
  \setlength{\itemsep}{0pt plus 0.5ex}
}
\newtheorem{theorem}{Lemma}
\definecolor{colorsrc}{rgb}{0.36, 0.54, 0.66}

\definecolor{colorlcksvd}{rgb}{0.91, 0.84, 0.42}
\definecolor{colorlcksvdd}{rgb}{0.8, 0.0, 0.1}
\definecolor{colorlcksvd}{rgb}{1, 0.56, 0.0}
\definecolor{colorfddl}{rgb}{0.44, 0.16, 0.39}
\definecolor{colordlsi}{rgb}{0.55, 0.71, 0.0}
\definecolor{colorcopar}{rgb}{0.9, .0, 0}
\definecolor{colorjdl}{rgb}{0.0, 0.55, 0.55}
\definecolor{colordlr}{rgb}{0.0, 0.55, 0.55}
\definecolor{colorlrsdl}{rgb}{0.0, 0.2, 1.0} 
\definecolor{colorlck}{rgb}{0.0, 0.9, 0.9}
\definecolor{pinegreen}{rgb}{0.0, 0.47, 0.44}

\def\x{{\mathbf x}}

\newcommand{\norm}[1]{\left\|#1\right\|}

\def\bmt{\left[\begin{matrix}}

\def\emt{\end{matrix}\right]}

\def\imply{\Rightarrow}

\def\mc {\mathcal}

\def\bx{\mathbf{x}}

\def\bd{\mathbf{d}}
\def\be{\mathbf{e}}
\def\fb{\mathbf{f}}

\def\bm{\mathbf{m}}
\def\M{\mathcal{M}}

\def\bu{\mathbf{u}}

\def\by{\mathbf{y}}

\def\and{\text{~and~}}

\def\trace{\textrm{trace}}
\def\etal{\textit{et al.}}
\def\R{\mathbb{R}}

\def\bzeros{\mathbf{0}}

\def\bA{\mathbf{A}}
\def\bB{\mathbf{B}}
\def\bD{\mathbf{D}}
\def\bE{\mathbf{E}}
\def\Fb{\mathbf{F}}
\def\bG{\mathbf{G}}

\def\bI{\mathbf{I}}

\def\bM{\mathbf{M}}

\def\bU{\mathbf{U}}
\def\bV{\mathbf{V}}
\def\bW{\mathbf{W}}
\def\bX{\mathbf{X}}

\def\bY{\mathbf{Y}}
\def\bZ{\mathbf{Z}}

\def\bW{\mathbf{W}}
\def\bw{\mathbf{w}}

\def\bDc{\bD_{0}}
\def\bXc{\bX^{0}}

\def\diag{\text{diag}}

\makeatletter
\newsavebox\myboxA
\newsavebox\myboxB
\newlength\mylenA
\newcommand*\lbar[2][.75]{%
    \sbox{\myboxA}{$\m@th#2$}%
    \setbox\myboxB\null
    \ht\myboxB=\ht\myboxA%
    \dp\myboxB=\dp\myboxA%
    \wd\myboxB=#1\wd\myboxA
    \sbox\myboxB{$\m@th\overline{\copy\myboxB}$}
    \setlength\mylenA{\the\wd\myboxA}
    \addtolength\mylenA{-\the\wd\myboxB}%
    \ifdim\wd\myboxB<\wd\myboxA%
       \rlap{\hskip 0.5\mylenA\usebox\myboxB}{\usebox\myboxA}%
    \else
        \hskip -0.3\mylenA\rlap{\usebox\myboxA}{\hskip 0.3\mylenA\usebox\myboxB}%
    \fi}
\makeatother

\def\lbD{\lbar{\bD}}
\def\lbY{\lbar{\bY}}
\def\lbX{\lbar{\bX}}
\def\R{\mathbb{R}}
\def\bY{\mathbf{Y}}
\usepackage[english]{babel}
\usepackage{fancyhdr}

\begin{document}
\title{Fast Low-rank Shared Dictionary Learning\\for Image Classification}
\author{Tiep Huu Vu,~\IEEEmembership{Student Member,~IEEE,}  Vishal Monga,~\IEEEmembership{Senior Member,~IEEE}
\thanks{The authors are with the School of Electrical Engineering and Computer Science,
The Pennsylvania State University, University Park, PA 16802, USA (e-mail: thv102@psu.edu).
\par
This work has been supported partially by the Office of Naval Research (ONR) under Grant 0401531 UP719Z0 and NSF CAREER award to (V.M.).}}
\maketitle

\begin{abstract}
\label{abstract}
Despite the fact that different objects possess distinct class-specific features, they also usually share common patterns. This observation has been exploited partially in a recently proposed dictionary learning framework by separating the particularity and the commonality (COPAR). Inspired by this, we propose a novel method to explicitly and simultaneously learn a set of common patterns as well as class-specific features for classification with more intuitive constraints. Our dictionary learning framework is hence characterized by both a shared dictionary and particular (class-specific) dictionaries. For the shared dictionary, we enforce a low-rank constraint, i.e. claim that its spanning subspace should have low dimension and the coefficients corresponding to this dictionary should be similar. For the particular dictionaries, we impose on them the well-known constraints stated in the Fisher discrimination dictionary learning (FDDL). Further, we develop new fast and accurate algorithms to solve the subproblems in the learning step, accelerating its convergence. The said algorithms could also be applied to FDDL and its extensions. The efficiencies of these algorithms are theoretically and experimentally verified by comparing their complexities and running time with those of other well-known dictionary learning methods. Experimental results on widely used image datasets establish the advantages of our method over state-of-the-art dictionary learning methods. 
\end{abstract}
\vspace{-.1in}
\section{Introduction}
\label{sec:intro}
Sparse representations have emerged as a powerful tool for a range of signal processing applications. Applications include {compressed sensing \cite{donoho2006compressed}, signal denoising, sparse signal recovery \cite{mousavi2015iterative}, image inpainting \cite{Aharon2006KSVD}, image segmentation \cite{spratling2013image}}, and more recently, signal classification. In such representations, most of signals can be expressed by a linear combination of few bases taken from a ``dictionary''. Based on this theory, a sparse representation-based classifier (SRC) \cite{Wright2009SRC} was initially developed for robust face recognition. Thereafter, SRC was adapted to numerous signal/image classification problems, ranging from medical image classification \cite{vu2015dfdl,vu2016tmi,Srinivas2014SHIRC}, hyperspectral image classification \cite{sun2015task,sun2014structured,chen2013hyperspectral}, synthetic aperture radar (SAR) image classification \cite{zhang2012multi}, recaptured image recognition \cite{thongkamwitoon2015image}, video anomaly detection \cite{mo2014adaptive}, and several others \cite{Mousavi2014ICIP,srinivas2015structured,zhang2012joint,dao2014structured,dao2016collaborative,van2013design,yang2010metaface,vu2016amp}.

\par
\begin{figure}[t]
\centering
\includegraphics[width = 0.49\textwidth]{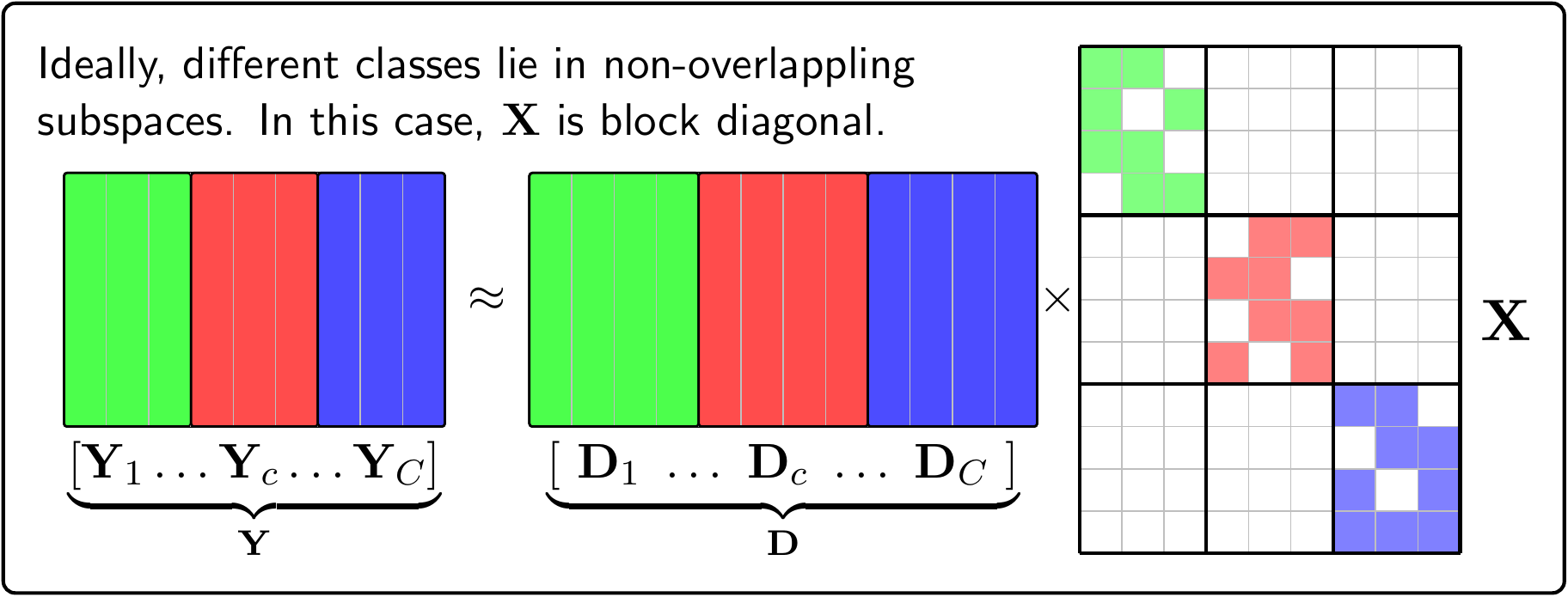}
\caption{\small Ideal structure of the coefficient matrix in SRC.}
\vspace{-.1in}
\label{fig:srcidea}

\end{figure}

The central idea in SRC is to represent a test sample (e.g. a face) as a linear combination of samples from the available training set. Sparsity manifests because most of non-zeros correspond to bases whose memberships are the same as the test sample. Therefore, in the ideal case, each object is expected to lie in its own class subspace and all class subspaces are non-overlapping. Concretely, given $C$ classes and a dictionary $\bD = [\bD_1, \dots, \bD_C]$ with $\bD_c$ comprising training samples from class $c, c = 1, \dots, C$, a new sample $\by$ from class $c$ can be represented as $\by \approx \bD_c\bx^c$. Consequently, if we express $\by$ using the dictionary $\bD: \by \approx \bD\bx = \bD_1\bx^1 + \dots + \bD_c\bx^c + \dots + \bD_C\bx^C$, then most of active elements of $\bx$ should be located in $\bx^c$ and hence, the coefficient vector $\x$ is expected to be sparse. In matrix form, let $\bY = [\bY_1, \dots, \bY_c, \dots, \bY_C]$ be the set of all samples where $\bY_c$ comprises those in class $c$, the coefficient matrix $\bX$ would be sparse. In the ideal case, $\bX$ is block diagonal (see Figure \ref{fig:srcidea}).
\pagestyle{headings}
 
\par
It has been shown that learning a dictionary from the training samples instead of using all of them as a dictionary can further enhance the performance of SRC. Most existing classification-oriented dictionary learning methods try to learn {\em discriminative class-specific dictionaries} by either imposing block-diagonal constraints on $\bX$ or encouraging the incoherence between class-specific dictionaries. Based on the K-SVD \cite{Aharon2006KSVD} model for general sparse representations, Discriminative K-SVD (D-KSVD) \cite{zhang2010discriminative} and Label-Consistent K-SVD (LC-KSVD) \cite{jiang2011learning,Zhuolin2013LCKSVD} learn the discriminative dictionaries by encouraging a projection of sparse codes $\bX$ to be close to a sparse matrix with all non-zeros being one while satisfying a block diagonal structure as in Figure \ref{fig:srcidea}. Vu \etal \cite{vu2015dfdl,vu2016tmi} with DFDL and Yang \etal~\cite{Meng2011FDDL,yang2014sparse} with FDDL apply Fisher-based ideas on dictionaries and sparse coefficients, respectively. Recently, Li \etal~\cite{li2014learning} with $D^2L^2R^2$ combined the Fisher-based idea and introduced a low-rank constraint on each sub-dictionary. They claim that such a model would reduce the negative effect of noise contained in training samples.
\vspace{-.3in}
\subsection{Closely Related work and Motivation} 
\vspace{-.1in}
\label{sub:subsection_name}
The assumption made by most discriminative dictionary learning methods, i.e. non-overlapping subspaces, is unrealistic in practice. Often objects from different classes share some common features, e.g. background in scene classification. This problem has been partially addressed by recent efforts, namely DLSI~\cite{ramirez2010classification}, COPAR~\cite{kong2012dictionary}, {JDL~\cite{zhou2014jointly}} and CSDL~\cite{gao2014learning}. However, DLSI does not explicitly learn shared features since they are still hidden in the sub-dictionaries. COPAR, JDL and CSDL explicitly learn a shared dictionary $\bD_0$ but suffer from the following drawbacks. First, we contend that the subspace spanned by columns of the shared dictionary must have low rank. Otherwise, class-specific features may also get represented by the shared dictionary. In the worst case, the shared dictionary span may include all classes, greatly diminishing the classification ability. Second, the coefficients (in each column of the sparse coefficient matrix) corresponding to the shared dictionary should be similar. This implies that features are shared between training samples from different classes via the ``shared dictionary''. In this paper, we develop a new low-rank shared dictionary learning framework (LRSDL) which satisfies the aforementioned properties. Our framework is basically a generalized version of the well-known FDDL \cite{Meng2011FDDL,yang2014sparse} with the additional capability of capturing shared features, resulting in better performance. We also show practical merits of enforcing these constraints are significant.
\par
The typical strategy in optimizing general dictionary learning problems is to alternatively solve their subproblems where sparse coefficients $\bX$ are found while fixing dictionary $\bD$ or vice versa. In discriminative dictionary learning models, both $\bX$ and $\bD$ matrices furthermore comprise of several small class-specific blocks constrained by complicated structures, usually resulting in high computational complexity. Traditionally, $\bX$, and $\bD$ are solved block-by-block until convergence. Particularly, each block $\bX_c$ (or $\bD_c$ in dictionary update ) is solved by again fixing all other blocks $\bX_i, i \neq c$ (or $\bD_i, i \neq c$). Although this greedy process leads to a simple algorithm, it not only produces inaccurate solutions but also requires huge computation. In this paper, we aim to mitigate these drawbacks by proposing efficient and accurate algorithms which allows to directly solve $\bX$ and $\bD$ in two fundamental discriminative dictionary learning methods: FDDL \cite{yang2014sparse} and DLSI \cite{ramirez2010classification}. These algorithms can also be applied to speed-up our proposed LRSDL, COPAR \cite{kong2012dictionary}, $D^2L^2R^2$ \cite{li2014learning} and other related works.
\vspace{-.3in}
\subsection{Contributions} 
\vspace{-.15in}
\label{sub:contributions}

The main contributions of this paper are as follows:
\noindent\begin{enumerate}
    \item {\bf A new low-rank shared dictionary learning framework}\footnote{The preliminary version of this work was presented in IEEE International Conference on Image Processing, 2016 \cite{vu2016icip}.} (LRSDL) for automatically extracting both discriminative and shared bases in several widely used image datasets is presented to enhance the classification performance of dictionary learning methods. Our framework simultaneously learns each class-dictionary per class to extract discriminative features and the shared features that all classes contain. {For the shared part, we impose two intuitive constraints. First, the shared dictionary must have a low-rank structure. Otherwise, the shared dictionary may also expand to contain discriminative features. Second, we contend that the sparse coefficients corresponding to the shared dictionary should be almost similar. In other words, the contribution of the shared dictionary to reconstruct every signal should be close together. We will experimentally show that both of these constraints are crucial for the shared dictionary.}

    \item {\bf New accurate and efficient algorithms for selected existing and proposed dictionary learning methods}. We present three effective algorithms for dictionary learning: i) sparse coefficient update in FDDL \cite{yang2014sparse} by using FISTA \cite{beck2009fast}. We address the main challenge in this algorithm -- how to calculate the gradient of a complicated function effectively -- by introducing a new simple function $\M(\bullet)$ on block matrices and a lemma to support the result. ii) Dictionary update in FDDL \cite{yang2014sparse} by a simple ODL \cite{mairal2010online} procedure using $\M(\bullet)$ and another lemma. {Because it is an extension of FDDL, the proposed LRSDL also benefits from the aforementioned efficient procedures.} iii) Dictionary update in DLSI \cite{ramirez2010classification} by a simple ADMM \cite{boyd2011distributed} procedure which requires only one matrix inversion instead of several matrix inversions as originally proposed in \cite{ramirez2010classification}. We subsequently show the proposed algorithms have both performance and computational benefits.

    \item {\bf Complexity analysis.} We derive the computational complexity of numerous dictionary learning methods in terms of approximate number of operations (multiplications) needed. We also report complexities and experimental running time of aforementioned efficient algorithms and their original counterparts.

    \item {\bf Reproducibility.} Numerous sparse coding and dictionary learning algorithms in the manuscript are reproducible via a user-friendly toolbox. The toolbox includes implementations of SRC \cite{Wright2009SRC}, ODL \cite{mairal2010online}, LC-KSVD \cite{Zhuolin2013LCKSVD}\footnote{Source code for LC-KSVD is directly taken from the paper at:\\ \texttt{http://www.umiacs.umd.edu/$\sim$zhuolin/projectlcksvd.html}.}, efficient DLSI \cite{ramirez2010classification}, efficient COPAR \cite{kong2012dictionary}, efficient FDDL \cite{yang2014sparse}, $D^2L^2R^2$ \cite{li2014learning} and the proposed LRSDL. The toolbox (a MATLAB version and a Python version) is provided\footnote{The toolbox can be downloaded at: \\{\texttt{http://signal.ee.psu.edu/lrsdl.html}}.} with the hope of usage in future research and comparisons via peer researchers.
\end{enumerate}

The remainder of this paper is organized as follows. Section \ref{sec:contribution} presents our proposed dictionary learning framework, the efficient algorithms for its subproblems and one efficient procedure for updating dictionaries in DLSI and COPAR. The complexity analysis of several well-known dictionary learning methods are included in Section \ref{sec:complexity_analysis}. In Section~\ref{sec:experiment_results}, we show classification accuracies of LRSDL on widely used datasets in comparisons with existing methods in the literature to reveal merits of the proposed LRSDL. Section \ref{sec:discussion_and_conclusion} concludes the paper.

\begin{figure}[t]
\centering
\includegraphics[width = 0.48\textwidth]{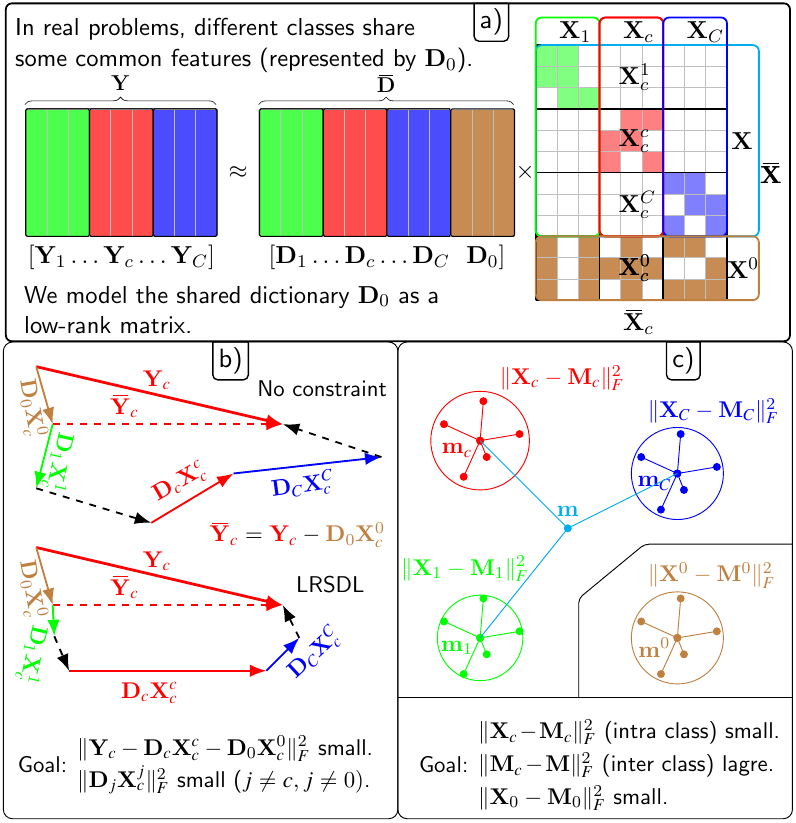}
\caption{\small {LRSDL idea with}: brown items -- shared; red, green, blue items
-- class-specific. a) Notation. b) The discriminative fidelity constraint: class-$c$ sample is mostly represented by $\bD_0$ and $\bD_c$. c) The Fisher-based discriminative coefficient constraint.}
\label{fig:lrsdl_idea}
\end{figure}

\section{Discriminative dictionary learning framework}
\label{sec:contribution}
\subsection{Notation} 
\label{sub:notaions}

In addition to notation stated in the Introduction, let $\bD_0$ be the shared dictionary, $\bI$ be the identity matrix with dimension inferred from context. For $c = 1, \dots, C$; $i = 0, 1, \dots, C$, suppose that $\bY_c \in \R^{d\times n_c}$ and $\bY \in \R^{d\times N}$ with $N = \sum_{c = 1}^C n_c$; $\bD_i \in \R^{d\times k_i}$, $\bD \in \R^{d\times K}$ with $K = \sum_{c=1}^C k_c$; and $\bX \in \R^{K\times N}$. Denote by $\bX^i$ the sparse coefficient of $\bY$ on $\bD_i$, by $\bX_c \in \R^{{K}\times N_c}$ the sparse coefficient of $\bY_c$ on ${\bD}$, by $\bX_c^i$ the sparse coefficient of $\bY_c$ on $\bD_i$. Let $\lbar{\bD} = \bmt\bD & \bD_0\emt$ be the total dictionary, $\lbar{\bX} = [\bX^T, (\bX^0)^T]^T$ and $\lbar{\bX}_c = [(\bX_c)^T, (\bX^0_c)^T ]^T$. For every dictionary learning problem, we implicitly constrain each basis to have its Euclidean norm no greater than 1. These variables are visualized in Figure \ref{fig:lrsdl_idea}a).
\par
Let $\bm, \bm^0,$ and $\bm_c$ be the mean of $\bX, \bX^0,$ and $\bX_c$ columns, respectively.
{Given a matrix $\mathbf{A}$ and a natural number $n$, define $\mu(\bA, n)$ as a matrix with $n$ same columns, each column being the mean vector of all columns of $\mathbf{A}$. If $n$ is ignored, we implicitly set $n$ as the number of columns of $\bA$.
 Let $\bM_c = \mu(\bX_c), \bM^0 = \mu(\bX^0)$, and $\bM = \mu(\bX, n)$ be the mean matrices. {The number of columns $n$ depends on context, e.g. by writing $\bM_c - \bM$, we mean that $n=n_c$} . The `mean vectors' are illustrated in Figure \ref{fig:lrsdl_idea}c).}

\par Given a function $f(A, B)$ with $A$ and $B$ being two sets of variables, define $f_A(B) = f(A, B)$ as a function of $B$ when the set of variables $A$ is fixed. Greek letters ($\lambda, \lambda_1, \lambda_2, \eta$) represent positive regularization parameters. Given a block matrix $\bA$, define a function $\M(\bA)$ as follows:
{ \begin{equation}
    \underbrace{\bmt
        \bA_{11} &  \dots & \bA_{1C}\\
        \bA_{21} &  \dots & \bA_{2C}\\
        \dots    &  \dots & \dots\\
        \bA_{C1} &  \dots & \bA_{CC}
        \emt}_{\bA}
    \mapsto \underbrace{\bA + \bmt
        \bA_{11} & \dots & \bzeros\\
        \bzeros  & \dots & \bzeros\\
        \dots    & \dots & \dots\\
        \bzeros  & \dots & \bA_{CC}
        \emt}_{\M(\bA)}.
\end{equation}}%
\noindent That is, $\M(\bA)$ doubles diagonal blocks of $\bA$. The row and column partitions of $\bA$ are inferred from context. $\M(\bA)$ is a computationally inexpensive function of $\bA$ and will be widely used in our LRSDL algorithm and the toolbox.
\par
We also recall here  the FISTA algorithm \cite{beck2009fast} for solving the family of problems:
\begin{equation}
\label{eqn:fista}
    \bX = \arg\min_{\bX} h(\bX) + \lambda\|\bX\|_1,
\end{equation}
where $h(\bX)$ is convex, continuously differentiable with Lipschitz continuous gradient. FISTA is an iterative method which requires to calculate gradient of $h(\bX)$ at each iteration. In this paper, we will focus on calculating the gradient of $h$.

\subsection{Closely related work: Fisher discrimination dictionary learning (FDDL)} 
\label{sub:fisher_discrimination_dictionary_learning}
FDDL \cite{Meng2011FDDL} has been used broadly as a technique for exploiting both structured dictionary and learning discriminative coefficient. Specifically, the discriminative dictionary $\bD$ and the sparse coefficient matrix $\bX$ are learned based on minimizing the following cost function:
\begin{eqnarray}
\label{eqn:fddl_cost_fn}
    J_{\bY}(\bD, \bX) = \frac{1}{2}f_{\bY}(\bD, \bX) + \lambda_1\|\bX\|_1 +
    \frac{\lambda_2}{2} g(\bX),
\end{eqnarray}
where  $\displaystyle f_{\bY}(\bD, \bX)  = \sum_{c=1}^C r_{\bY_c}(\bD, \bX_c)$ is the discriminative fidelity with:\\
\scalebox{.97}{\parbox{\linewidth}{%
\begin{equation*}
r_{\bY_c}(\bD, \bX_c) = \|\bY_c - \bD\bX_c\|_F^2 +  \|\bY_c - \bD_c\bX_c^c\|_F^2 + \sum_{j\neq c}\|\bD_j\bX^j_c\|_F^2,
\end{equation*}%
}}
$ g(\bX) = \sum_{c=1}^C (\|\bX_c - \bM_c\|_F^2 - \|\bM_c - \bM\|_F^2) + \|\bX\|_F^2$ is the Fisher-based discriminative coefficient term, and the $l_1$-norm encouraging the sparsity of coefficients.
{The last term in $r_{\bY_c}(\bD, \bX_c)$ means that $\bD_j$ has a  small contribution to the representation of $\bY_c$ for all $j \neq c$. With the last term $\|\bX\|_F^2$ in $g(\bX)$, the cost function becomes convex with respect to $\bX$.}

\begin{figure}[t]
\centering
\includegraphics[width = 0.48\textwidth]{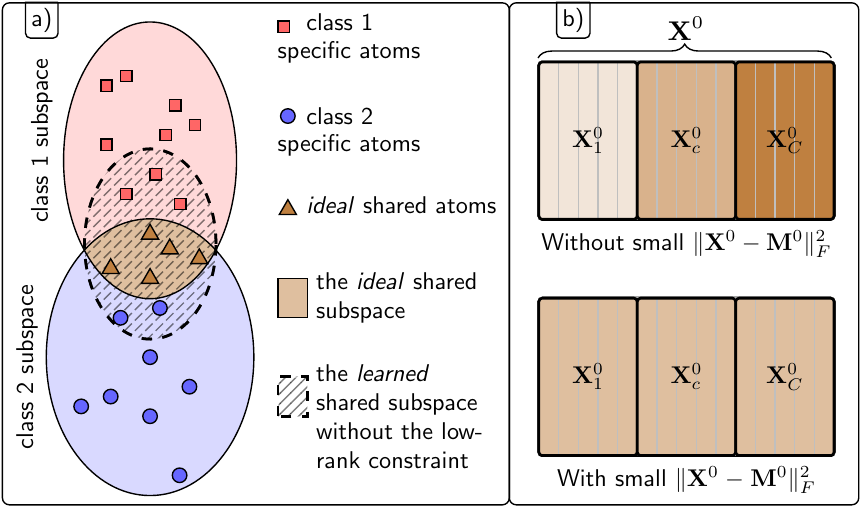}
\caption{\small Two constraints on the shared dictionary. a) Low-rank constraints. b) Similar shared codes.}
\label{fig:whylowrank}
\end{figure}

\vspace{-.2in}
{
\subsection{Proposed Low-rank shared dictionary learning (LRSDL) }
\label{sub:low_rand_shared_dictionary_learning_}
The shared dictionary needs to satisfy the following properties: \\
\textbf{\textit{1) Generativity:}}
As the common part, the most important property of the shared dictionary is to represent samples from all classes \cite{kong2012dictionary,zhou2014jointly,gao2014learning}. In other words, it is expected that $\bY_c$ can be well represented by the collaboration of the particular dictionary $\bD_c$ and the shared dictionary $\bD_0$. Concretely, the discriminative fidelity term $f_{\bY}(\bD, \bX)$ in (\ref{eqn:fddl_cost_fn}) can be extended to $\lbar{f}_{\bY}(\lbar{\bD}, \lbar{\bX}) = \sum_{c=1}^C \lbar{r}_{\bY_c} (\lbD, \lbX_c)$ with $\lbar{r}_{\bY_c} (\lbD, \lbX_c)$ being defined as:
\begin{equation*}
\|\bY_c - \lbar{\bD}\lbar{\bX}_c\|_F^2 + \|\bY_c - \bD_c\bX_c^c -
\bD_0\bX^0_c\|_F^2 + \sum_{j = 1, j\neq c}^C \|\bD_j\bX^j_c\|_F^2.
\end{equation*}
Note that since $\bar{r}_{\bY_c}(\lbD, \lbX_c) = r_{\bar{\bY}_c}(\bD, \bX_c)$ with $\lbar{\bY}_c = \bY_c - \bD_0\bX^0_c$ (see Figure \ref{fig:lrsdl_idea}b)), we have:
\begin{equation*}
\label{eqn:f_fbar}
\lbar{f}_{\bY}(\lbD, \lbX) = f_{\bar{\bY}}(\bD, \bX),
\end{equation*}
with $\lbar{\bY} = \bY - \bD_0\bX^0.$\\
This generativity property can also be seen in Figure \ref{fig:whylowrank}a). In this figure, the intersection of different subspaces, each representing one class, is one subspace visualized by the light brown region. One class subspace, for instance class 1, can be well represented by the \textit{ideal} shared atoms (dark brown triangles) and the corresponding class-specific atoms (red squares).}
\par 
{
\textbf{\textit{2) Low-rankness:}}
The stated generativity property  is only the necessary condition for a set of atoms to qualify a shared dictionary. Note that the set of atoms inside the shaded ellipse in Figure \ref{fig:whylowrank}a) also satisfies the generativity property: along with the \textit{remaining} red squares, these atoms well represent class 1 subspace; same can be observed for class 2. In the worst case, the set including all the atoms can also satisfy the generativity property, and in that undesirable case, there would be no discriminative features remaining in the class-specific dictionaries.  Low-rankness is hence {\em necessary} to prevent the shared dictionary from \textit{absorbing} discriminative atoms. The constraint is natural based on the observation that the subspace spanned by the shared dictionary has low dimension. Concretely, we use the nuclear norm regularization $\|\bD_0\|_*$, which is the convex relaxation of $\text{rank}(\bD_0)$ \cite{recht2010guaranteed}, to force the shared dictionary to be low-rank. In contrast with our work, existing approaches that employ shared dictionaries, i.e.\ COPAR~\cite{kong2012dictionary} and JDL~\cite{zhou2014jointly}, do not incorporate this crucial constraint.
}

\par 
{
\textbf{\textit{3) Code similarity:}}
In the classification step, a test sample $\by$ is decomposed into two parts: the part represented by the shared dictionary $\bD_0\bx^0$ and the part expressed by the remaining dictionary $\bD\bx$. Because $\bD_0\bx^0$ is not expected to contain class-specific features, it can be excluded before doing classification. The shared code $\bx^0$ can be considered as the contribution of the shared dictionary to the representation of $\by$. Even if the shared dictionary already has low-rank, its contributions to each class might be different as illustrated in Figure \ref{fig:whylowrank}b), the top row. In this case, the different contributions measured by $\bX^0$ convey class-specific features, which we aim to avoid. Naturally, the regularization term $\|\bX^0 - \bM^0\|$ is added to our proposed objective function to force each $\bx^0$ to be close to the mean vector $\bm^0$ of all $\bX^0$. \\
With this constraint, the Fisher-based discriminative coefficient term $g(\bX)$ is extended to $\lbar{g}(\lbar{\bX})$ defined as:
\begin{equation}
\label{eqn:deffbar}
\lbar{g}(\lbar{\bX}) = g(\bX) + \|\bX^0 -\bM^0\|_F^2,
\end{equation}
Altogether, the cost function $\lbar{J}_{\bY}(\lbar{\bD}, \lbar{\bX})$ of our proposed LRSDL is:
\begin{equation}
\label{eqn:lrsdl_cost_fn}
\lbar{J}_{\bY}(\lbD, \lbX) = \frac{1}{2}\lbar{f}_{\bY}(\lbar{\bD}, \lbar{\bX})
+ \lambda_1\|\lbar{\bX}\|_1 + \frac{\lambda_2}{2} \lbar{g}(\lbar{\bX})
+ \eta \|\bD^0\|_*.
\end{equation}
By minimizing this objective function, we can jointly find the class specific and shared dictionaries. Notice that if there is no shared dictionary $\bD_0$ (by setting $k_0 = 0$), then $\lbar{\bD}, \lbar{\bX}$ become $\bD, \bX$, respectively, $\lbar{J}_{\bY}(\lbar{\bD}, \lbar{\bX}) $ becomes $J_{\bY}(\bD, \bX)$ and LRSDL reduces to FDDL.
}

\par
\noindent \textbf{Classification scheme:}
\par
After the learning process, we obtain the total dictionary $\lbar{\bD}$ and mean vectors $\bm_c, \bm^0$. For a new test sample $\by$, first we find its coefficient vector $\lbar{\bx} = [\bx^T, (\bx^0)^T]^T$ with the sparsity constraint on $\lbar{\bx}$ and further encourage $\bx^0$ to be close to $\bm^0$:
\begin{equation}
\label{eqn:findcodebarx}
    \lbar{\bx} = \arg\min_{\lbar{\bx}} \frac{1}{2}\|\by - \lbar{\bD}\lbar{\bx}\|_2^2
    + \frac{\lambda_2}{2}\|\bx^0 - \bm^0\|_2^2 + \lambda_1\|\lbar{\bx}\|_1.
\end{equation}

Using $\lbar{\bx}$ as calculated above, we extract the contribution of the shared dictionary to obtain $\lbar{\by} = \by - \bD_0\bx^0$. The identity of $\by$ is determined by:
\begin{equation}
    \arg\min_{1 \leq c \leq C} (w\|\lbar{\by} - \bD_c\bx^c\|_2^2
    + (1-w)\|\bx - \bm_c\|_2^2),
\end{equation}
where $w \in [0,1]$ is a preset weight for balancing the contribution of the two terms.
\subsection{Efficient solutions for optimization problems} 
\label{sub:solving_the_opt}
Before diving into minimizing the LRSDL objective function in (\ref{eqn:lrsdl_cost_fn}),
 we first present efficient algorithms for minimizing the FDDL objective
 function in (\ref{eqn:fddl_cost_fn}).

\subsubsection{Efficient FDDL dictionary update} 
\label{ssub:fddl_dictionary_update}
\par
Recall that in \cite{Meng2011FDDL}, the dictionary update step is divided into $C$ subproblems, each updates one class-specific dictionary $\bD_c$ while others fixed. This process is repeated until convergence. This approach is not only highly time consuming but also inaccurate. We will see this in a small example presented in Section \ref{sub:valid_eff_algs}. We refer this original FDDL dictionary update as O-FDDL-D.

\par We propose here an efficient algorithm for updating dictionary called E-FDDL-D where the \textit{total dictionary} $\bD$ will be optimized when $\bX$ is fixed, significantly reducing the computational cost.
\par {Concretely, when we fix $\bX$ in equation \eqref{eqn:fddl_cost_fn}, the problem of solving $\bD$ becomes}:
\begin{equation}
\label{eqn:fddl_updateD}
    \bD = \arg\min_{\bD}f_{\bY, \bX}(\bD)
\end{equation}
\vspace{-.1in}
Therefore, $\bD$ can be solved by using the following lemma.
\begin{theorem}
\label{lem:fddl_updateD}
\textit{The optimization problem (\ref{eqn:fddl_updateD}) is equivalent to: }
\begin{equation}
    \label{eqn:efddl_d_odl}
    \bD = \arg\min_{\bD}\{ -2\trace(\bE\bD^T) + \trace(\Fb\bD^T\bD) \},
\end{equation}
\textit{where $\bE = \bY\M(\bX^T)$ and $\Fb = \M(\bX\bX^T)$. }
\end{theorem}
\textit{Proof:} See Appendix \ref{apd:proof_fddl_updateD}.
\par
The problem \eqref{eqn:efddl_d_odl} can be solved effectively by Online Dictionary Learning (ODL) method
\cite{mairal2010online}.

\subsubsection{Efficient FDDL sparse coefficient update (E-FDDL-X)} 
\label{ssub:eff_fddl_sparse_coefficient_update}
When $\bD$ is fixed, $\bX$ will be found by solving:
\begin{equation}
\label{eqn:fddl_updateX}
    \bX = \arg\min_{\bX} h(\bX)    + \lambda_1 \|\bX\|_1,
\end{equation}
where $h(\bX) = \frac{1}{2} f_{\bY, \bD}(\bX)  + \frac{\lambda_2}{2}g(\bX)$.
The problem (\ref{eqn:fddl_updateX}) has the form of equation (\ref{eqn:fista}), and can hence be solved by FISTA \cite{beck2009fast}. We need to calculate gradient of
$f(\bullet)$ and $g(\bullet)$ with respect to $\bX$.
\begin{theorem}
\label{lem:fddl_updateX}
{\textit{ Calculating gradient of $h(\bX)$ in equation \eqref{lem:fddl_updateX} }}
    \begin{eqnarray}
    \frac{\partial \frac{1}{2}f_{\bY, \bD}(\bX)}{\partial \bX} =& \M(\bD^T\bD) \bX - \M(\bD^T\bY), \\
     \frac{\partial\frac{1}{2}g(\bX)}{\partial(\bX)} =& 2\bX + \bM -
    2\underbrace{\bmt \bM_1 & \bM_2 & \dots \bM_c\emt}_{\widehat{\bM}}.
\end{eqnarray}
    Then we obtain:
{\small \begin{equation}
    \label{eqn:fddl_dif_x}
    \frac{\partial h(\bX)}{\partial \bX} =
        (\M(\bD^T\bD) + 2\lambda_2 \bI) \bX - \M(\bD^T\bY) + \lambda_2(\bM - 2\widehat{\bM}).
\end{equation}}
\end{theorem}
\textit{Proof:} See Appendix \ref{apd:proof_fddl_updateX}.
\par {Since the proposed LRSDL is an extension of FDDL, we can also extend these two above algorithms to optimize LRSDL cost function as follows. }

\subsubsection{LRSDL dictionary update (LRSDL-D)} 
\label{ssub:lrsdl_dictionary_update}
Returning to our proposed LRSDL problem, we need to find $\lbD = [\bD,\bD_0]$ when $\lbX$ is fixed. We propose a method to solve $\bD$ and $\bD_0$ separately.
\par
\textbf{For updating $\bD$}, recall the observation that $\lbar{f}_{\bY}(\lbD, \lbX) = f_{\lbar{\bY}}(\bD, \bX)$, with $\lbY \triangleq \bY - \bD_0\bX^0$ (see equation \eqref{eqn:f_fbar}), and the E-FDDL-D presented in section \ref{ssub:fddl_dictionary_update}, we have:
\begin{equation}
    \bD = \arg\min_{\bD}\{-2\trace(\bE\bD^T) + \trace(\Fb\bD^T\bD)\},
\end{equation}
with $\bE = \lbY\M(\bX^T)$ and $\Fb = \M(\bX\bX^T).$
\par
\textbf{For updating $\bD_0$}, we use the following lemma:
\begin{theorem}
\label{lem:lrsdld0x0}
{\textit{When $\bD, \bX$ in \eqref{eqn:lrsdl_cost_fn} are fixed,}}
\begin{eqnarray}
    \nonumber
    \lbar{J}_{\bY, \bD, \bX}(\bD_0, \bX_0) &=& \|\bV - \bD_0\bX_0\|_F^2 + \frac{\lambda_2}{2}\|\bX^0 - \bM^0\|_F^2 + \\
    \label{eqn:lrsdl_updateD0}
    && + \eta\|\bD_0\|_* + \lambda_1\|\bX^0\|_1 + \text{constant},
\end{eqnarray}
where $\bV = \bY - \frac{1}{2}\bD\M(\bX)$.
\end{theorem}
\textit{Proof:} See Appendix \ref{apd:proof_lrsdl_updateD0X0}.
\par
Based on the Lemma \ref{lem:lrsdld0x0}, $\bD_0$ can be updated by solving:
\begin{align}
\nonumber
    &\bD_0 = \arg\min_{\bD_0} \trace{(\Fb\bD_0^T\bD_0)} -2\trace(\bE\bD_0^T) +  \eta\|\bD_0\|_* \\
    \label{eqn:lrsdl_d0_form_EF}
    &\text{where:}~~\bE = \bV(\bX^0)^T; \quad\quad\Fb = \bX^0(\bX^0)^T
\end{align}
using the ADMM \cite{boyd2011distributed} method and the singular value thresholding algorithm \cite{cai2010singular}. The ADMM procedure is as follows. First, we choose a positive $\rho$, initialize $\bZ= \bU = \bD_0$,
then alternatively solve each of the following subproblems until convergence:
\begin{align}
\label{eqn:lrsdl_d0_1}
    \bD_0 =& \arg\min_{\bD_0} -2\trace(\lbar{\bE}\bD_0^T) + \trace\left(\lbar{\Fb}\bD_0^T\bD_0\right),\\
\label{eqn:lrsdl_d0_2}
     & \text{with~~} \lbar{\bE} = \bE + \frac{\rho}{2} (\bZ - \bU); ~ \lbar{\Fb} = \Fb + \frac{\rho}{2} \bI,\\
\label{eqn:lrsdl_d0_3}
    \bZ =& \mathcal{D}_{\eta/\rho}(\bD_c + \bU),\\
\label{eqn:lrsdl_d0_4}
    \bU =& \bU + \bD_0 - \bZ,
\end{align}

where $\mathcal{D}$ is the shrinkage thresholding operator \cite{cai2010singular}. {The optimization problem \eqref{eqn:lrsdl_d0_1} can be solved by ODL \cite{mairal2010online}. Note that \eqref{eqn:lrsdl_d0_2} and \eqref{eqn:lrsdl_d0_4} are computationally inexpensive. }


\vspace{-.1in}
\subsubsection{LRSDL sparse coefficients update (LRSDL-X)} 
\label{sub:sparse_coding_update}
In our  preliminary work \cite{vu2016icip}, we proposed a method for effectively solving $\bX$ and
$\bX^0$ alternatively, now we combine both problems into one and find
$\lbX$ by solving the following optimization problem:
\begin{equation}
    \lbX = \arg\min_{\lbX} \lbar{h}(\lbX) + \lambda_1\|\lbX\|_1.
\end{equation}
where $\displaystyle \lbar{h}(\lbX) = \frac{1}{2}\lbar{f}_{\bY, \overline{\bD}}(\lbX) + \frac{\lambda_2}{2} \lbar{g}(\lbX)$.
We again solve this problem using FISTA \cite{beck2009fast} with the gradient of $\displaystyle \lbar{h}(\lbX)$:
\begin{equation}
\displaystyle
    \label{eqn:hbar_diff_both}
    \nabla \lbar{h}(\lbX) = 
    \bmt \displaystyle \frac{\partial \lbar{h}_{\bX^0}(\bX)}{\partial \bX}\\
         \displaystyle \frac{\partial \lbar{h}_{\bX}(\bX^0)}{\partial \bX^0} \emt.
\end{equation}
\textit{For the upper term}, by combining the observation
\begin{eqnarray}
\nonumber
 \lbar{h}_{\bX^0}(\bX) &=& \frac{1}{2}\lbar{f}_{\bY, \overline{\bD}, \bX^0}(\bX) +
                            \frac{\lambda_2}{2} \lbar{g}_{\bX^0}(\bX) , \\
 &=& \frac{1}{2}f_{\lbar{\bY}, \bD}(\bX) + \frac{\lambda_2}{2}g(\bX) + \text{constant},
\end{eqnarray}
  and using equation, we obtain:
    \begin{equation}
    \label{eqn:hbar_diff_up}
        \displaystyle \frac{\partial \lbar{h}_{\bX^0}(\bX)}{\partial \bX} =
                (\M(\bD^T\bD) + 2\lambda_2 \bI) \bX - \M(\bD^T\lbY) + \lambda_2(\bM - 2\widehat{\bM}).
    \end{equation}

\textit{For the lower term}, by using Lemma \ref{lem:lrsdld0x0}, we have:
    \begin{equation}
        \lbar{h}_{\bX}(\bX^0) = \|\bV - \bD_0\bX^0\|_F^2 + \frac{\lambda_2}{2}\|\bX^0 - \bM^0\|_F^2 + \text{constant}.
    \end{equation}
    \begin{eqnarray}
        \nonumber
         \imply \frac{\partial \lbar{h}_{\bX}(\bX^0)}{\partial \bX^0} =
         2\bD_0^T\bD_0\bX^0 - 2\bD_0^T\bV + \lambda_2(\bX^0 - \bM^0),\\
         \label{eqn:hbar_diff_lo}
         = (2\bD_0^T\bD_0 + \lambda_2\bI)\bX^0 - 2\bD_0^T\bV -\lambda_2\bM^0.
    \end{eqnarray}

{By combining these two terms, we can calculate \eqref{eqn:hbar_diff_both}.}
\par 
{
Having $\nabla \lbar{h}(\lbX)$ calculated, we can update $\lbX$ by the FISTA algorithm \cite{beck2009fast} as given in Algorithm 1. Note that we need to compute a Lipschitz coefficient $L$ of $\nabla \lbar{h}(\lbX)$. 
The overall algorithm of LRSDL is given in Algorithm 2.
}
\begin{algorithm}[t]
\label{alg:LRSDLX}
    \caption{{LRSDL sparse coefficients update by FISTA\cite{beck2009fast}}}
    \begin{spacing}{1.3}
    \begin{algorithmic}
    \Function {$(\hat{\bX}, \hat{\bX}^0)$ = LRSDL\_X}{$\bY, \bD, \bD_0, \bX, \bX^0, \lambda_1, \lambda_2$}.
    \State 1. Calculate:
    \begin{align*}
        \mathbf{A} &= \mathcal{M}(\bD^T\bD) + 2\lambda_2 \mathbf{I}; \\
        \mathbf{B} &= 2\bD_0^T\bD_0 + \lambda_2\mathbf{I} \\
        L &= \lambda_{\max}(\mathbf{A}) + \lambda_{\max}({\mathbf{B}}) + 4\lambda_2 + 1 \text{\footnotemark}
    \end{align*}
    \State 2. Initialize $\bW_1 = \bZ_0 = \bmt \bX \\ \bX^0\emt, t_1 = 1, k = 1$
    \While {\text{not convergence and $ k < k_{\max}$}}
        \State 3. Extract $\bX, \bX^0$ from $\bW_k$.
        \State 4. Calculate gradient of two parts:
        \begin{eqnarray*}
            \bM &=& \mu(\bX), \bM_c = \mu(\bX_c), \widehat{\bM} = [\bM_1, \dots, \bM_C].\\
            \bV &=& \bY - \frac{1}{2}\bD\mathcal{M}(\mathbf{X})\\
            \bG &=& \bmt
                \bA\bX - \mathcal{M}(\bD^T(\bY - \bD_0\bX^0)) + \lambda_2(\bM - \widehat{\bM})\\
                \mathbf{B}\bX^0 - \bD_0^T\bV -\lambda_2\mu(\bX^0)
            \emt
        \end{eqnarray*}
        \State 5. $\bZ_k = \mathcal{S}_{\lambda_1/L}\left(\bW_k - \bG/L\right)$ ($\mathcal{S}_{\alpha}()$ is the element-wise soft thresholding function. $\mathcal{S}_{\alpha}(x) = \text{sgn}(x)(|x| - \alpha)_+$).

        \State 6. $t_{k+1} = (1 + \sqrt{1 + 4t_k^2})/2$

        \State 7. $\bW_{k+1} = \bZ_k + \frac{t_k - 1}{t_{k+1}} (\bZ_k - \bZ_{k-1})$
        \State 8. $k = k + 1$
    \EndWhile
    \State 9. OUTPUT: Extract $\bX, \bX^0$ from $\bZ_k$.
    \EndFunction
    \end{algorithmic}
    \end{spacing}
\end{algorithm}
\footnotetext{In our experiments, we practically choose this value as an upper bound of the Lipschitz constant of the gradient.}

\begin{algorithm}[t!]
    \caption{{LRSDL algorithm}}
    \begin{spacing}{1.2}
    \begin{algorithmic}

    \Function {$(\hat{\bX}, \hat{\bX}^0)$ = LRSDL}{$\bY, \lambda_1, \lambda_2, \eta$}.
    \State 1. Initialization $\bX = \mathbf{0}$, and:
    \begin{eqnarray*}
        (\bD_c, \bX_c^c) &=&\arg\min_{\bD, \bX} \frac{1}{2} \|\bY_c - \bD\bX\|_F^2 + \lambda_1 \|\bX\|_1\\
        (\bD_0, \bX^0) &=&\arg\min_{\bD, \bX} \frac{1}{2} \|\bY - \bD\bX\|_F^2 + \lambda_1 \|\bX\|_1
    \end{eqnarray*}
    \While {not converge}
        \State 2. Update $\bX$ and $\bX^0$ by Algorithm 1.
        \State 3. Update $\bD$ by ODL \cite{mairal2010online}:
        \begin{eqnarray*}
            \mathbf{E} &=& (\bY - \bD_0\bX^0)\mathcal{M}(\bX^T)\\
            \Fb &=& \mathcal{M}(\bX\bX^T) \\
            \bD &=& \arg\min_{\bD}\{-2\trace(\mathbf{E}\bD^T) + \trace(\Fb\bD^T\bD)\}
        \end{eqnarray*}
        \State 4. Update $\bD_0$ by ODL \cite{mairal2010online} and ADMM \cite{boyd2011distributed} (see equations \eqref{eqn:lrsdl_d0_1} - \eqref{eqn:lrsdl_d0_4}).
    \EndWhile

    \EndFunction

    \end{algorithmic}
    \end{spacing}
\end{algorithm}

\subsection{Efficient solutions for other dictionary learning methods} 
\vspace{-.1in}
\label{sub:edlsi}
We also propose here another efficient algorithm for updating dictionary in
two other well-known dictionary learning methods: DLSI \cite{ramirez2010classification}
and COPAR \cite{kong2012dictionary}. 
\par
The cost function $J_1(\bD, \bX)$
in DLSI is defined as:
\begin{equation}
\label{eqn:dlsi_cost_function}
    \sum_{c = 1}^C\big(||\bY_c - \bD_c \bX^c\|_F^2 +
    \lambda\|\bX^c\|_1 +       \frac{\eta}{2}\sum_{j=1,j \neq c}^C \|\bD_j^T\bD_c\|_F^2   \big)
\end{equation}
Each class-specific dictionary $\bD_c$ is updated by fixing others and solve:
\begin{equation}
\label{eqn:dlsi_updateDc}
    \bD_c = \arg\min_{\bD_c} \|\bY_c - \bD_c\bX^c\|_F^2 + \eta\|\bA\bD_c\|_F^2,
\end{equation}
with $\bA = \bmt \bD_1, \dots , \bD_{c-1}, \bD_{c+1}, \dots, \bD_{C}\emt^T$.
\par
The original solution for this problem, which will be referred as O-FDDL-D,
updates each column $\bd_{c,j}$ of $\bD_c$ one by one based on the procedure:
\begin{align}
    \label{eqn:odlsid_step1}
    \bu &= (\|\bx_c^j\|_2^2 \bI + \eta\bA^T\bA)^{-1}(\bY_c - \displaystyle\sum_{i \neq j}\bd_{c, i}\bx_c^i)\bx_c^j,\\
    \label{eqn:odlsid_step2}
    \bd_{c, j} &= \displaystyle \bu/\|\bu\|_2^2,
\end{align}

where $\bd_{c, i}$ is the $i$-th column of $\bD_c$ and $\bx_c^j$ is the $j$-th row of $\bX_c$. This algorithm is highly computational since it requires one matrix inversion for \textit{each} of $k_c$ columns of $\bD_c$. We propose one ADMM \cite{boyd2011distributed} procedure to update $\bD_c$ which requires \textit{only one} matrix inversion, which will be referred as E-DLSI-D. First, by letting $\bE = \bY_c(\bX^c)^T$ and $\Fb = \bX^c(\bX^c)^T$, we rewrite (\ref{eqn:dlsi_updateDc}) in a more general form:
\begin{equation}
    \label{eqn:dlsi_updateDc_admm_form}
    \bD_c = \displaystyle\arg\min_{\bD_c}\trace(\Fb\bD_c^T\bD_c) -2\trace(\bE\bD_c^T) + \eta \|\bA\bD_c\|_F^2.
\end{equation}

In order to solve this problem, first, we choose a $\rho$, let $\bZ= \bU = \bD_c$,
then alternatively solve each of the following sub problems until convergence:
\begin{align}
\label{eqn:edlsid_firststep}
    \bD_c =& \arg\min_{\bD_c} -2\trace(\lbar{\bE}\bD_c^T) + \trace\left(\lbar{\Fb}\bD_c^T\bD_c\right),\\
\label{eqn:edlsid_2}
     & \text{with~~} \lbar{\bE} = \bE + \frac{\rho}{2} (\bZ - \bU); ~ \lbar{\Fb} = \Fb + \frac{\rho}{2} \bI.\\
\label{eqn:edlsid_3}
    \bZ =& {(2\eta \bA^T\bA + \rho \bI)^{-1}}(\bD_c + \bU). \\
\label{eqn:edlsid_laststep}
    \bU =& \bU + \bD_c - \bZ.
\end{align}
This efficient algorithm requires only one matrix inversion. Later in this paper,
we will both theoretically and experimentally show that E-DLSI-D is much more
efficient than O-DLSI-D \cite{ramirez2010classification}. Note that this algorithm can be beneficial for two
subproblems of updating the common dictionary and the particular dictionary
in COPAR \cite{kong2012dictionary} as well.
\section{Complexity analysis} 
\label{sec:complexity_analysis}
We compare the computational complexity for the efficient algorithms and their corresponding original algorithms. We also evaluate the total complexity of the proposed LRSDL and competing dictionary learning methods: DLSI \cite{ramirez2010classification}, COPAR \cite{kong2012dictionary} and FDDL \cite{Meng2011FDDL}. The complexity for each algorithm is estimated as the (approximate) number of multiplications required for one iteration (sparse code update and dictionary update). For simplicity, we assume: i) number of training samples, number of dictionary bases in each class (and the shared class) are the same, which means: $n_c = n, k_i = k$. ii) The number of bases in each dictionary is comparable to number of training samples per class and much less than the signal dimension, i.e. $k \approx n \ll d$. iii) Each iterative algorithm requires $q$ iterations to convergence. For consistency, we have changed notations in those methods by denoting $\bY$ as training sample and $\bX$ as the sparse code.

\par
In the following analysis, we use the fact that: i) if $\bA \in \R^{m\times n}, \bB \in \R^{n \times p}$, then the matrix multiplication $\bA\bB$ has complexity $mnp$. ii) If $\bA \in \R^{n \times n}$ is nonsingular, then the matrix inversion $\bA^{-1}$ has complexity $n^3$. iii) The singular value decomposition of a matrix $\bA\in \R^{p \times q}$, $p > q$, is assumed to have complexity $O(pq^2)$.
\subsection{Online Dictionary Learning (ODL)} 
\vspace{-.1in}
\label{sub:online_dictionary_learning}
We start with the well-known Online Dictionary Learning \cite{mairal2010online} whose cost function is:
\begin{equation}
    J(\bD, \bX) = \frac{1}{2} \|\bY - \bD\bX\|_F^2 + \lambda\|\bX\|_1.
\end{equation}
where $\bY \in \R^{d \times n}, \bD \in \R^{d \times k}, \bX \in \R^{k \times n}$.
Most of dictionary learning methods find their solutions by alternatively solving one variable while fixing others. There are two subproblems:
\subsubsection{Update $\bX$ (ODL-X)}
\label{ssub:odl_x}%
When the dictionary $\bD$ is fixed, the sparse coefficient $\bX$ is updated by solving the problem:
\begin{equation}
\label{eqn:l1norm}
    \bX = \arg\min_{\bX} \frac{1}{2} \|\bY - \bD\bX\|_F^2 + \lambda \|\bX\|_1
\end{equation}
using FISTA \cite{beck2009fast}. In each of $q$ iterations, the most computational task is to compute $\bD^T\bD \bX - \bD^T\bY$ where $\bD^T\bD$ and $\bD^T\bY$ are precomputed with complexities $k^2d$ and $kdn$, respectively. The matrix multiplication $(\bD^T\bD)\bX$ has complexity $k^2n$. Then, the total complexity of ODL-X is:
\begin{equation}
\label{eqn:cplxt_l1norm}
    k^2d + kdn + qk^2n = k(kd + dn + qkn).
\end{equation}
\subsubsection{Update $\bD$ (ODL-D)}
After finding $\bX$, the dictionary $\bD$ will be updated by:
\begin{align}
    \bD &= \arg\min_{\bD} -2\trace(\bE\bD^T) + \trace(\Fb\bD^T\bD),
\end{align}
   subject to: $\|\bd_i\|_2 \leq 1$, with $\bE = \bY\bX^T$, and $\Fb = \bX\bX^T$.\\
Each column of $\bD$ will be updated by fixing all others:
\begin{eqnarray*}
    \label{eqn:odl_u}
     \bu \leftarrow \frac{1}{\Fb_{ii}} (\be_i - \bD \fb_i) - \bd_i;~~\bd_i & \leftarrow & \frac{\bu}{\max(1, \|\bu\|_2)},
\end{eqnarray*}
where $\bd_i, \be_i, \fb_i$ are the $i-$th columns of $\bD, \bE, \Fb$ and $\Fb_{ii}$ is the $i-$th element in the diagonal of $\Fb$. The dominant computational task is to compute $\bD\fb_i$ which requires $dk$ operators. Since $\bD$ has $k$ columns and the algorithm requires $q$ iterations,
the complexity of ODL-D is $qdk^2$.
\subsection{Dictionary learning with structured incoherence (DLSI)} 
\label{sub:dictionary_update_in_dlsi}
DLSI \cite{ramirez2010classification} proposed a method to encourage the independence
between bases of different classes by minimizing coherence between cross-class bases. The cost function $J_1(\bD, \bX)$ of DLSI is defined as \eqref{eqn:dlsi_cost_function}.
\subsubsection{Update $\bX$ (DLSI-X)}
In each iteration, the algorithm solves $C$ subproblems:
\begin{equation}
    \bX^c = \arg\min_{\bX^c} \|\bY_c - \bD_c\bX^c\|_F^2 + \lambda\|\bX^c\|_1.
\end{equation}
with $\bY_c \in \R^{d\times n}, \bD_c \in \R^{d \times k}$, and $\bX^c \in \R^{k \times n}$. Based on~\eqref{eqn:cplxt_l1norm}, the complexity of updating $\bX$ ($C$ subproblems) is:
\begin{equation}
    Ck(kd + dn + qkn).
\end{equation}
\subsubsection{Original update $\bD$ (O-DLSI-D)}
For updating $\bD$, each sub-dictionary $\bD_c$ is solved via (\ref{eqn:dlsi_updateDc}).
The main step in the algorithm is stated in (\ref{eqn:odlsid_step1}) and (\ref{eqn:odlsid_step2}).
    The dominant computational part is the matrix inversion which has complexity $d^3$. Matrix-vector multiplication and vector normalization can be ignored here.
    Since $\bD_c$ has $k$ columns, and the algorithm requires $q$ iterations,
    the complexity of the O-DLSI-D algorithm is $Cqkd^3$.
\subsubsection{Efficient update $\bD$ (E-DLSI-D)} 
Main steps of the proposed algorithm are presented in equations (\ref{eqn:edlsid_firststep})--(\ref{eqn:edlsid_laststep}) where (\ref{eqn:edlsid_2}) and (\ref{eqn:edlsid_laststep}) require much less computation compared to (\ref{eqn:edlsid_firststep}) and (\ref{eqn:edlsid_3}). The total (estimated) complexity of efficient $\bD_c$ update is a summation of two terms: i) $q$ times ($q$ iterations) of ODL-D in (\ref{eqn:edlsid_firststep}). ii) One matrix inversion ($d^3$) and $q$ matrix multiplications in (\ref{eqn:edlsid_3}). Finally, the complexity of E-DLSI-D is:
\begin{equation}
 C(q^2dk^2 + d^3 + qd^2k) = Cd^3 + Cqdk(qk+d).
\end{equation}
Total complexities of O-DLSI (the combination of DLSI-X and O-DLSI-D) and E-DLSI (the combination of DLSI-X and E-DLSI-D) are summarized in Table \ref{tab:complexity_analysis}.

\begin{table}[t]
\centering
\caption{Complexity analysis for proposed efficient algorithms and their original versions}
\label{tab:complexity_analysis_subproblems}
{\small \begin{tabular}{|l|l|l|}
\hline
Method & Complexity &\begin{tabular}[c]{@{}c@{}} Plugging \\ numbers\end{tabular}\\
\hline
O-DLSI-D & $Cqkd^3$                  & $6.25\times 10^{12}$\\
E-DLSI-D & $Cd^3 + Cqdk(qk + k)$     & $2.52\times 10^{10}$\\
\hline
O-FDDL-X & $C^2k(dn + qCkn + Cdk)$   & $1.51 \times 10^{11}$\\
E-FDDL-X & $C^2k(dn + qCnk + dk)$    & $1.01 \times 10^{11}$\\
\hline
O-FDDL-D & $Cdk(qk + C^2n)$          & $10^{11}$\\
E-FDDL-D & $Cdk(Cn + Cqk) + C^3k^2n$ & $2.8\times 10^{10}$\\
\hline
\end{tabular}}
\end{table}

\vspace{-.26in}
\subsection{Separating the particularity and the commonality dictionary learning (COPAR)} 
\label{sub:COPAR}
\subsubsection{Cost function}
{COPAR \cite{kong2012dictionary} is another dictionary learning method which also considers the shared dictionary (but without the low-rank constraint)}.
By using the same notation as in LRSDL, we can rewrite the cost function of COPAR in the following form:
\begin{align*}
\frac{1}{2}f_1(\bY, \lbD, \lbX) + \lambda\norm{\lbX}_1 +
                \eta\sum_{c=0}^C\sum_{i=0, i \neq c}^C\|\bD_i^T\bD_c\|_F^2,
\end{align*}
where $\displaystyle f_1(\bY, \lbD, \lbX) = \sum_{c=1}^Cr_1(\bY_c, \lbD, \lbX_c)$ and $r_1(\bY_c, \lbD, \lbX_c)$ is defined as:
\begin{align*}
     \|\bY_c - \lbD \lbX_c\|_F^2 + \|\bY_c - \bD_0\bX^0_c - \bD_c \bX^c_c\|_F^2 + \sum_{j=1,j\neq c}^C \|\bX_c^j\|_F^2.
\end{align*}

\subsubsection{Update $\bX$ (COPAR-X)}
In sparse coefficient update step, COPAR \cite{kong2012dictionary} solve $\lbX_c$ one by one via one $l_1$-norm regularization problem:
\begin{equation*}
    \tilde{\bX} = \arg\min_{\tilde{\bX}} \|\tilde{\bY} - \tilde{\bD}\tilde{\bX}\|_F^2 + \tilde{\lambda}\|\tilde{\bX}\|_1,
\end{equation*}
where $\tilde{\bY} \in \R^{\tilde{d} \times n}, \tilde{\bD} \in \R^{\tilde{d} \times \tilde{k}}$, $ \tilde{\bX} \in \R^{(\tilde{k} \times n}$, $\tilde{d} =2d + (C-1)k$ and $\tilde{k} = (C+1)k$ (details can be found in Section 3.1 of \cite{kong2012dictionary}). Following results in Section \ref{ssub:odl_x} and supposing that $C \gg 1, q \gg 1, n \approx k \ll d$, the complexity of COPAR-X is:
\begin{eqnarray*}
    C\tilde{k}(\tilde{k}\tilde{d} + \tilde{d}n + q\tilde{k}n) &\approx& C^3k^2(2d + Ck + qn).
\end{eqnarray*}
\vspace{-.3in}
\subsubsection{Update $\bD$ (COPAR-D)}
The COPAR dictionary update algorithm requires to solve $(C+1)$ problems of form (\ref{eqn:dlsi_updateDc_admm_form}). While O-COPAR-D uses the same method as O-DLSI-D (see equations (\ref{eqn:odlsid_step1}-\ref{eqn:odlsid_step2})), the proposed E-COPAR-D takes advantages of E-DLSI-D presented in Section \ref{sub:edlsi}. Therefore, the total complexity of O-COPAR-D is roughly $Cqkd^3$, while the total complexity of E-COPAR-D is roughly $C(q^2dk^2 + d^3 + qd^2k).$ {Here we have supposed $C+1 \approx C$ for large $C$}.

Total complexities of O-COPAR (the combination of COPAR-X and O-COPAR-D) and E-COPAR (the combination of COPAR-X and E-COPAR-D) are summarized in Table \ref{tab:complexity_analysis}.

\subsection{Fisher discrimination dictionary learning (FDDL)} 
\vspace{-.1in}
\label{sub:sparse_coding_update_in_fddl}
\subsubsection{Original update $\bX$ (O-FDDL-X)}
Based on results reported in DFDL~\cite{vu2016tmi}, the complexity of O-FDDL-X is roughly $C^2kn(d + qCk) + C^3dk^2 = C^2k(dn + qCkn + Cdk)$.

\subsubsection{Efficient update $\bX$ (E-FDDL-X)}
Based on section \ref{ssub:eff_fddl_sparse_coefficient_update}, the complexity of E-FDDL-X mainly comes from equation (\ref{eqn:fddl_dif_x}). Recall that function $\M(\bullet)$ does not require much computation. The computation of $\bM$ and ${\bM_c}$ can also be neglected since each required calculation of one column, all other columns are the same. Then the total complexity of the algorithm E-FDDL-X is roughly:
\begin{align}
\nonumber
    &\underbrace{(Ck)d(Ck)}_{\M(\bD^T\bD + \lambda_2\bI)}  +
    \underbrace{(Ck)d(Cn)}_{\M(\bD^T\bY)} +
    q\underbrace{(Ck)(Ck)(Cn)}_{\M(\bD^T\bD + \lambda_2\bI)\bX}, \\
    &=   C^2k(dk + dn + qCnk).
\end{align}

\subsubsection{Original update $\bD$ (O-FDDL-D)}
The original dictionary update in FDDL is divided in to $C$ subproblems. In each subproblem, one dictionary $\bD_c$ will be solved while all others are fixed via:
\par{\small\begin{align}
    \nonumber
    \bD_c &= \arg\min_{\bD_c} \|\widehat{\bY} - \bD_c \bX^c\|_F^2 + \|\bY_c - \bD_c\bX^c_c\|_F^2  + \sum_{i \neq c} \|\bD_c\bX^c_i\|_F^2, \\
        &= \underbrace{\arg\min_{\bD_c} -2\trace(\bE \bD_c^T) + \trace(\Fb\bD_c^T\bD_c)}_{\text{complexity: ~} qdk^2},
\end{align}}
\\where:
\begin{align*}
    \widehat{\bY} &= \bY - \sum_{i \neq c} \bD_i\bX^i & \text{complexity:}~ (C-1)dkCn,\\
    \bE &= {\widehat{\bY}(\bX^c)^T} + \bY_c (\bX_c^c)^T & \text{complexity:}~ d(Cn)k + dnk, \\
    \Fb &=  2(\bX^c)(\bX^c)^T& \text{complexity}~ k(Cn)k.
\end{align*}
When $d \gg k, C \gg 1$, complexity of updating $\bD_c$ is:
\begin{equation}
    qdk^2 + (C^2 + 1)dkn + Ck^2n \approx qdk^2 + C^2dkn
\end{equation}
Then, complexity of O-FDDL-D is $Cdk(qk + C^2n)$.


\subsubsection{Efficient update $\bD$ (E-FDDL-D)}
Based on Lemma \ref{lem:fddl_updateD}, the complexity of E-FDDL-D is:
\begin{align}
\nonumber
    &\underbrace{d(Cn)(Ck)}_{\bY\M(\bX)^T} +
    \underbrace{(Ck)(Cn)(Ck)}_{\M(\bX\bX^T)} +
    \underbrace{qd(Ck)^2}_{\text{ODL in (\ref{eqn:efddl_d_odl})}},  \\
    & = Cdk(Cn + Cqk) + C^3k^2n.
\end{align}
Total complexities of O-FDDL and E-FDDL are summarized in Table \ref{tab:complexity_analysis}.


\subsection{LRSDL} 
\label{sub:lrsdl}
\subsubsection{Update $\bX, \bX^0$}
From (\ref{eqn:hbar_diff_both}), (\ref{eqn:hbar_diff_up}) and (\ref{eqn:hbar_diff_lo}), in each iteration of updating $\lbX$, we need to compute:
\begin{eqnarray*}
    &&(\M(\bD^T\bD) + 2\lambda_2\bI)\bX - {\M(\bD^T\bY)} +\\
    &&\quad\quad\quad +\lambda_2(\bM - 2\widehat{\bM}) - \M(\bD^T\bD_0\bX^0), ~\text{and} \\
    &&{(2\bD_0^T\bD_0 + \lambda_2\bI)}\bX^0 - {2\bD_0^T\bY} + \bD_0^T\bD\M(\bX) - \lambda_2 \bM^0.
\end{eqnarray*}
Therefore, the complexity of LRSDL-X is:
\begin{eqnarray}
\nonumber
    &&\underbrace{(Ck)d(Ck)}_{\bD^T\bD} +
    \underbrace{(Ck)d(Cn)}_{\bD^T\bY} +
    \underbrace{(Ck)dk}_{\bD^T\bD_0} +
    \underbrace{kdk}_{\bD_0^T\bD_0} +
    \underbrace{kd(Cn)}_{\bD_0^T\bY} + \\
\nonumber
    &&+ q\left(
    \begin{matrix}
    \underbrace{(Ck)^2(Cn)}_{(\M(\bD^T\bD) + 2\lambda_2\bI)\bX } +
    \underbrace{(Ck)k(Cn)}_{\M(\bD^T\bD_0\bX^0)} + \\
    +\underbrace{k^2Cn}_{{(2\bD_0^T\bD_0 + \lambda_2\bI)}\bX^0} +
    \underbrace{k(Ck)(Cn)}_{ \bD_0^T\bD\M(\bX)}
    \end{matrix}
    \right), \\
\nonumber
\label{eqn:complexity_lrsdl_X}
    &&\approx C^2k(dk + dn) + Cdk^2 + qCk^2n(C^2 + 2C + 1),\\
    && \approx C^2k(dk + dn + qCkn).
\end{eqnarray}
which is similar to the complexity of E-FDDL-X.
Recall that we have supposed number of classes $C \gg 1$.

\subsubsection{Update $\bD$}
Compare to E-FDDL-D, LRSDL-D requires one more computation of $\lbY = \bY - \bD_0\bX^0$ (see section \ref{ssub:lrsdl_dictionary_update}). Then, the complexity of LRSDL-D is:
\begin{align}
\nonumber
    &\underbrace{Cdk(Cn + Cqk) + C^3k^2n}_{\text{E-FDDL-D}} + \underbrace{dk(Cn)}_{\bD_0\bX^0}, \\
    \label{eqn:complexity_lrsdl_D}
    &\approx Cdk(Cn + Cqk) + C^3k^2n,
\end{align}
which is similar to the complexity of E-FDDL-D.
\subsubsection{Update $\bD_0$}
The algorithm of LRSDL-D0 is presented in section \ref{ssub:lrsdl_dictionary_update} with the main computation comes from (\ref{eqn:lrsdl_d0_form_EF}), (\ref{eqn:lrsdl_d0_1}) and (\ref{eqn:lrsdl_d0_3}). The shrinkage thresholding operator in (\ref{eqn:lrsdl_d0_3}) requires one SVD and two matrix multiplications. The total complexity of LRSDL-D0 is:
\begin{align}
\nonumber
    &\underbrace{d(Ck)(Cn)}_{\bV = \bY - \frac{1}{2} \bD\M(\bX)} +
    \underbrace{d(Cn)k}_{\bE \text{~in (\ref{eqn:lrsdl_d0_form_EF})}} +
    \underbrace{k(Cn)k}_{\Fb \text{~in (\ref{eqn:lrsdl_d0_form_EF})}} +
    \underbrace{qdk^2}_{(\ref{eqn:lrsdl_d0_1})} +
    \underbrace{O(dk^2) + 2dk^2}_{(\ref{eqn:lrsdl_d0_3})}, \\
    \nonumber
    &\approx C^2dkn + qdk^2 + O(dk^2),\\
    \label{eqn:complexity_LRSDL_D0}
    & = C^2dkn + (q + q_2)dk^2, ~~\text{for some}~ q_2.
\end{align}
{By combing \eqref{eqn:complexity_lrsdl_X}, \eqref{eqn:complexity_lrsdl_D} and \eqref{eqn:complexity_LRSDL_D0}, we obtain the total complexity of LRSDL, which is specified in the last row of Table \ref{tab:complexity_analysis}.}
\def\ct{$\pm$}
\begin{table}[t]
\caption{Complexity analysis for different dictionary learning methods}
    \centering
    \label{tab:complexity_analysis}
    {\small
    \begin{tabular}{|l|l|l|}
    \hline
    Method & Complexity & \begin{tabular}[c]{@{}c@{}} Plugging \\ numbers\end{tabular}\\ \hline
    \hline
    O-DLSI & $Ck(kd + dn + qkn)+ Cqkd^3 $& $6.25\times 10^{12}$\\
    \hline
    E-DLSI & \begin{tabular}[c]{@{}c@{}} $Ck(kd + dn + qkn) +$ \\ $Cd^3 + Cqdk(qk+d) $ \end{tabular}& $3.75\times 10^{10}$\\
    \hline
    O-FDDL  & \begin{tabular}[c]{@{}c@{}} $C^2dk(n+Ck+Cn) +$ \\ $+Ck^2q(d + C^2n)$ \end{tabular}& $2.51\times 10^{11}$\\
    \hline
    E-FDDL & $ C^2k((q+1)k(d + Cn) + 2dn)    $  & $1.29\times 10^{11}$\\ 
    \hline
    O-COPAR & $C^3k^2(2d + Ck + qn) + Cqkd^3 $  & $6.55\times 10^{12}$\\
    \hline
    E-COPAR & \begin{tabular}[c]{@{}c@{}}$C^3k^2(2d + Ck + qn)+$ \\ $+Cd^3 + Cqdk(qk+d)$\\ \end{tabular}  & $3.38\times 10^{11}$\\
    \hline
    LRSDL & \begin{tabular}[c]{@{}c@{}}$C^2k((q+1)k(d + Cn) + 2dn)$ \\ $C^2dkn + (q+q_2)dk^2 $ \end{tabular} & $1.3\times 10^{11}$\\
    \hline
    \end{tabular}
    }
\end{table}

\subsection{Summary}
\label{sub:complexity_analysis_summary}
Table \ref{tab:complexity_analysis_subproblems} and Table \ref{tab:complexity_analysis} show final complexity analysis of each  proposed efficient algorithm and their original counterparts. Table \ref{tab:complexity_analysis} compares LRSDL to other state-of-the-art methods. We pick a typical set of parameters with 100 classes, 20 training samples per class, 10 bases per sub-dictionary and shared dictionary, data dimension 500 and 50 iterations for each iterative method. Concretely, $C = 100,~ n = 20,~ k = 10,~ q = 50,~ d = 500$. We also assume that in (\ref{eqn:complexity_LRSDL_D0}), $q_2 = 50$. Table \ref{tab:complexity_analysis_subproblems} shows that all three proposed efficient algorithms require less computation than original versions with most significant improvements for speeding up DLSI-D. Table \ref{tab:complexity_analysis} demonstrates an interesting fact. LRSDL is the least expensive computationally when compared with other \textit{original} dictionary learning algorithms, and only E-FDDL has lower complexity, which is to be expected since the FDDL cost function is a special case of the LRSDL cost function. COPAR is found to be the most expensive computationally.

\section{Experimental results} 
\label{sec:experiment_results}
\vspace{-0.1in}
\subsection{Comparing methods and datasets}
\label{sub:methods_datasets}
\vspace{-0.15in}
\begin{figure}[t]
\centering
\includegraphics[width = 0.45\textwidth]{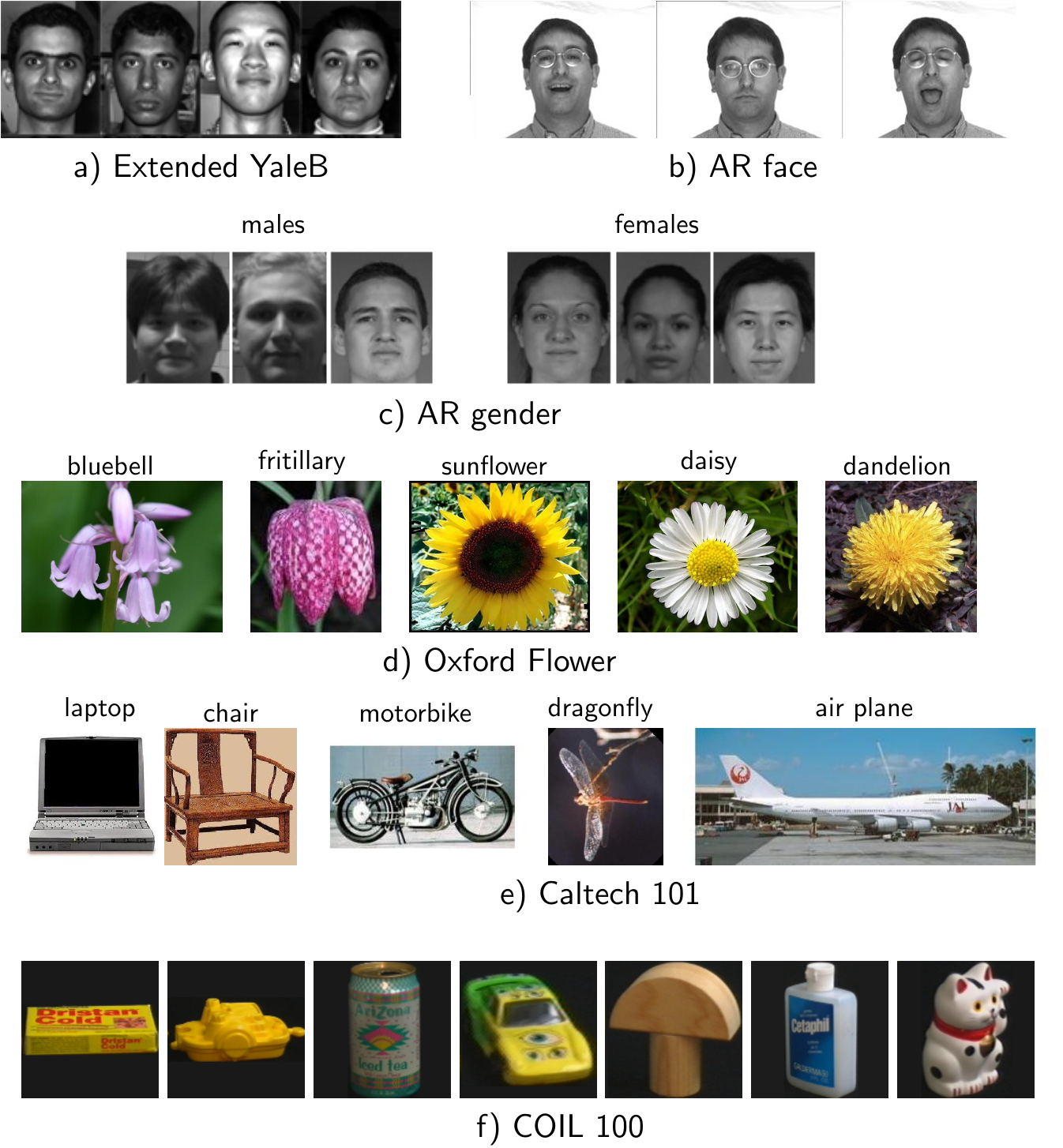}
\caption{\small {Examples from six datasets}. }
\label{fig:examples}
\end{figure}

We present the experimental results of applying these methods to five diverse
datasets: the Extended YaleB face dataset \cite{georghiades2001few}, the AR
face dataset~\cite{ardataset}, the AR gender dataset, the Oxford Flower dataset~\cite{Nilsback06}, and two multi-class object category dataset --
the Caltech 101~\cite{fei2007learning} {and COIL-100~\cite{nene1996columbia}}. Example images from these datasets are
shown in Figure \ref{fig:examples}. We compare our results with those using
SRC~\cite{Wright2009SRC} and other state-of-the-art dictionary learning methods:
LC-KSVD~\cite{Zhuolin2013LCKSVD}, DLSI~\cite{ramirez2010classification},
FDDL~\cite{yang2014sparse}, COPAR~\cite{kong2012dictionary}, $D^2L^2R^2$~\cite{li2014learning}, {DLRD~\cite{ma2012sparse}, JDL~\cite{zhou2014jointly}, and SRRS~\cite{li2016learning}}.
{Regularization parameters in all methods are chosen using five-fold cross-validation \cite{Kohavi95astudy}.} {For each experiment, we use 10 different randomly split training and test sets and report averaged results.}

\begin{figure}[t]
\centering
\includegraphics[width = 0.48\textwidth]{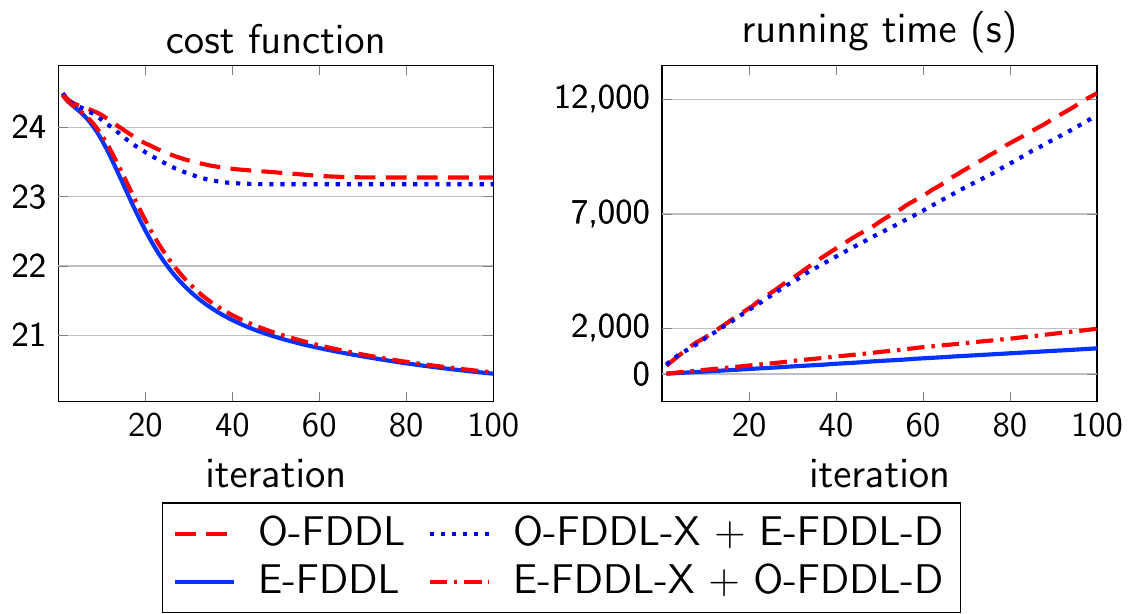}
\caption{\small Original and efficient FDDL convergence rate comparison. }
\label{fig:compare_fddl_algs}
\end{figure}

\begin{figure}[t]
\centering
\includegraphics[width = 0.48\textwidth]{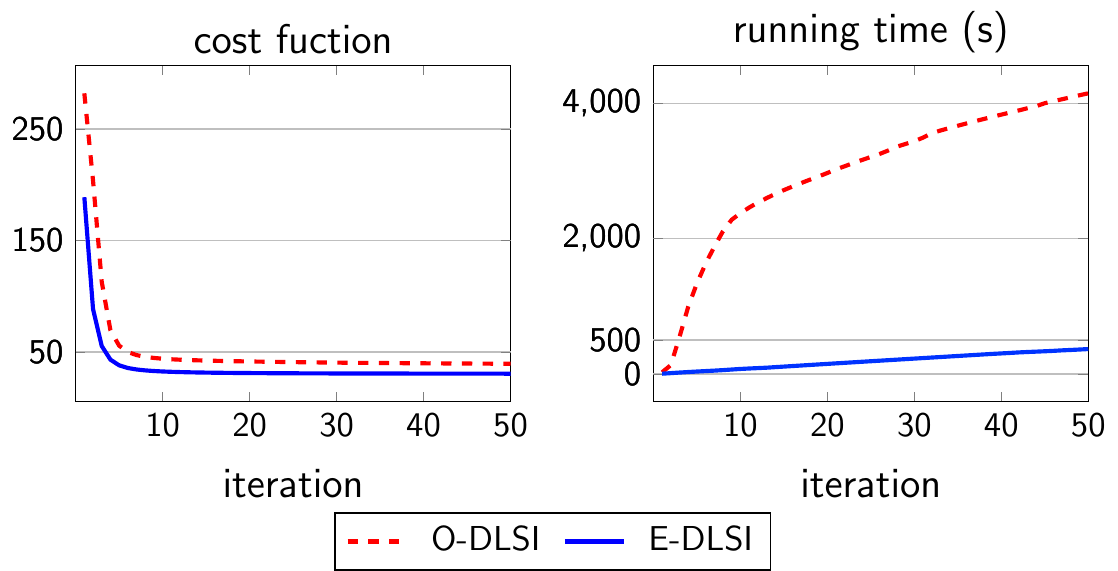}
\caption{\small DLSI convergence rate comparison. }
\label{fig:compare_dlsi_alg}
\end{figure}

\begin{figure}[t]
\centering
\includegraphics[width = 0.48\textwidth]{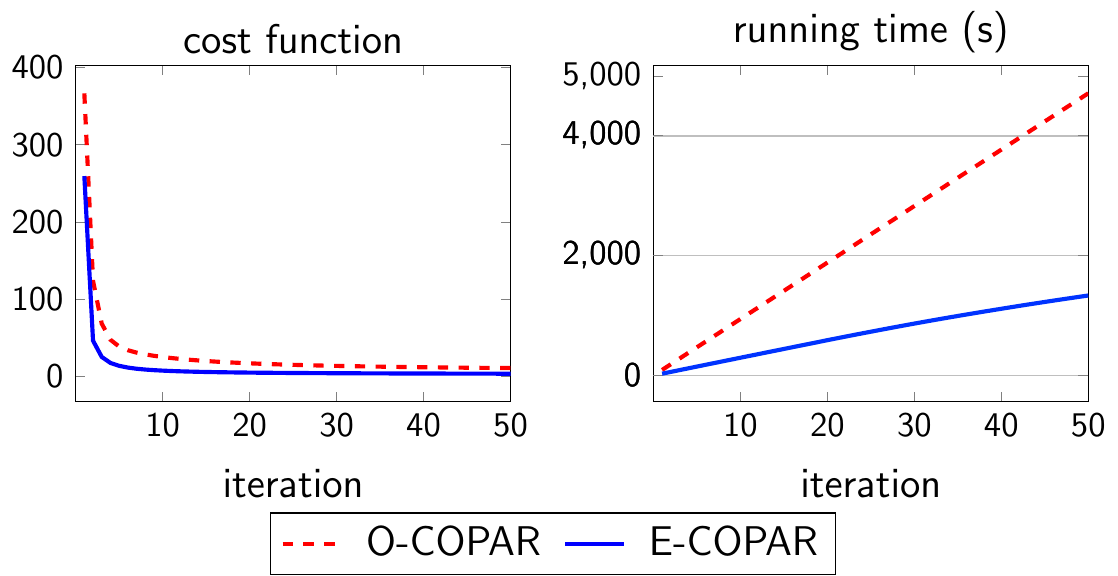}
\caption{\small COPAR convergence rate comparison.}
\label{fig:compare_COPAR_alg}
\end{figure}

\par For {\bf two face datasets}, feature descriptors are random faces, which are made by  projecting a face image onto a random vector using a random projection matrix.  As in
\cite{zhang2010discriminative}, the dimension of a random-face feature in the
Extended YaleB is $d= 504$, while the dimension in AR face is $d = 540$. Samples of these two datasets are shown in  Figure \ref{fig:examples}a) and b).

\par For {\bf the AR gender dataset}, we first choose a non-occluded subset (14 images per person)
from the AR face dataset, which consists of 50 males and 50 females, to conduct experiment of gender
classification. Training images are taken from the first 25 males and 25 females, while test images
comprises all samples from the remaining 25 males and 25 females. PCA was used to reduce the dimension
of each image to 300. Samples of this dataset are shown in Figure \ref{fig:examples}c).
\par {\bf The Oxford Flower dataset} is a collection of images of flowers drawn from 17 species with
80 images per class, totaling 1360 images. For feature extraction, based on the impressive results
presented in \cite{yang2014sparse}, we choose the Frequent Local Histogram feature
extractor \cite{fernando2012effective} to obtain feature vectors of dimension 10,000.
The test set consists of 20 images per class, the remaining 60 images per class are
used for training. Samples of this dataset are shown in Figure \ref{fig:examples}d).

\par For the {\bf Caltech 101 dataset}, we use a dense SIFT (DSIFT) descriptor. The DSIFT
descriptor is extracted from $25\times 25$ patch which is densely sampled on a dense grid
with 8 pixels. We then extract the sparse coding spatial pyramid matching (ScSPM) feature \cite{yang2009linear},
which is the concatenation of vectors pooled from words of the extracted DSIFT descriptor.
Dimension of words is 1024 and max pooling technique is used with pooling grid of
$1\times 1, 2 \times 2$, and $4 \times 4$. With this setup, the dimension of ScSPM
feature is 21,504; this is followed by dimension reduction to $d = 3000$ using PCA.

\begin{figure*}[t]
\centering
  \includegraphics[width=\textwidth]{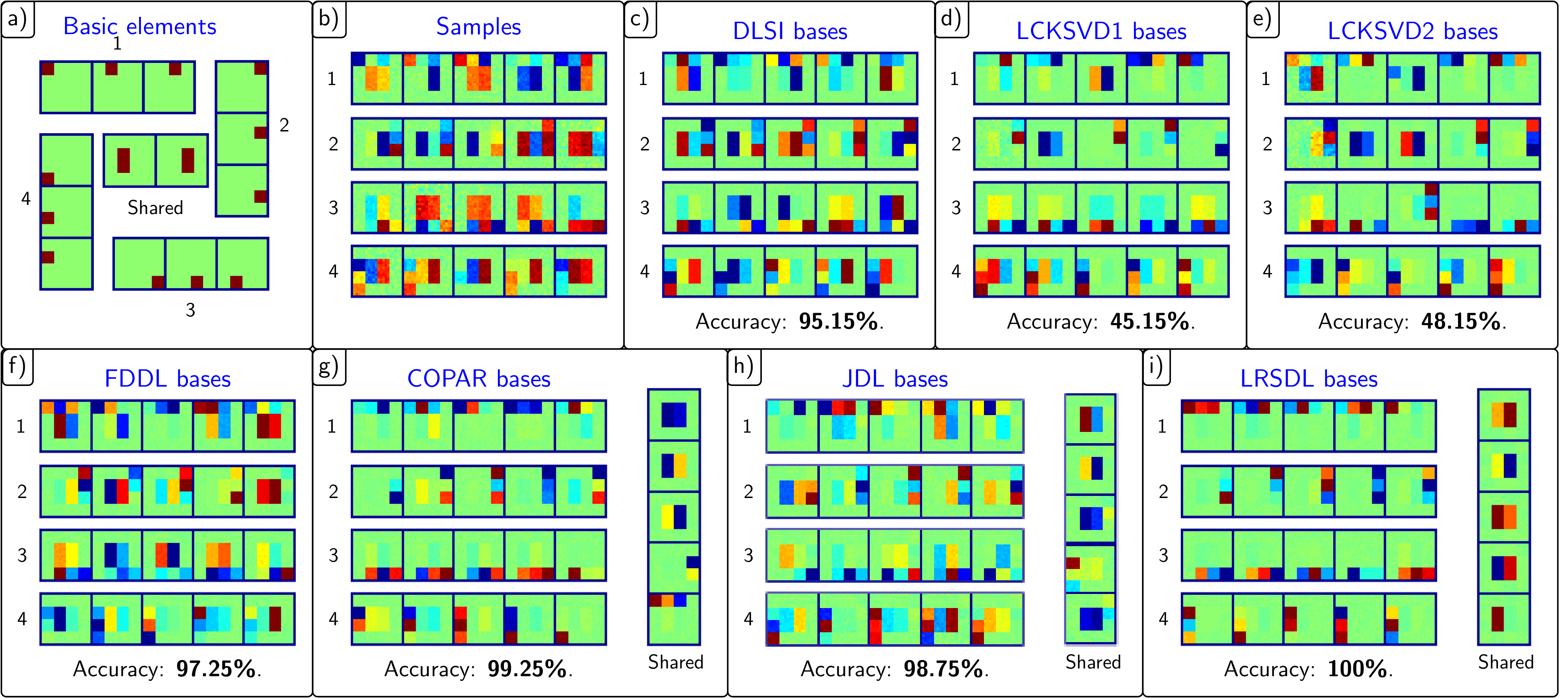}
  \vspace{-0.1in}
 \caption{\small Visualization of learned bases of different dictionary learning methods on the simulated data.}
  \label{fig: bases_simulated}
\end{figure*}

\par {The {\bf COIL-100 dataset} contains various views of 100 objects with different lighting conditions. Each object has 72 images captured from equally spaced views. Similar to the work in~\cite{li2016learning}, we randomly choose 10 views of each object for training, the rest is used for test. To obtain the feature vector of each image, we first convert it to grayscale, resize to 32 $\times$ 32 pixel, vectorize this matrix to a 1024-dimensional vector, and finally normalize it to have unit norm. }

\par
Samples of this dataset are shown in Figure \ref{fig:examples}e).
\vspace{-0.2in}
\subsection{Validation of efficient algorithms} 
\label{sub:efficient_algs}
\vspace{-0.1in}
\label{sub:valid_eff_algs}


To evaluate the improvement of three efficient algorithms proposed in section \ref{sec:contribution}, we apply these efficient algorithms and their original versions on training samples from the AR face dataset to verify the convergence speed of those algorithms. In this example, number of classes $C = 100$, the random-face
feature dimension $d = 300$, number of training samples per class $n_c = n = 7$,
number of atoms in each particular dictionary $k_c = 7$.
\subsubsection{E-FDDL-D and E-FDDL-X}
Figure \ref{fig:compare_fddl_algs} shows the cost functions and running time after each of 100 iterations of 4 different versions of FDDL: the original FDDL (O-FDDL), combination of O-FDDL-X and E-FDDL-D, combination of E-FDDL-X and O-FDDL-D, and the efficient FDDL (E-FDDL). The first observation is that O-FDDL converges quickly to a suboptimal solution, which is far from the best cost obtained by E-FDDL. In addition, while O-FDDL requires more than 12,000 seconds (around 3 hours and 20 minutes) to run 100 iterations, it takes E-FDDL only half an hour to do the same task.

\subsubsection{E-DLSI-D and E-COPAR-D}

Figure \ref{fig:compare_dlsi_alg} and \ref{fig:compare_COPAR_alg} compare convergence rates of DLSI and COPAR algorithms. As we can see, while the cost function value improves slightly, the run time of efficient algorithms reduces significantly. Based on benefits in both cost function value and computation, in the rest of this paper, we use efficient optimization algorithms instead of original versions for obtaining classification results. 

\def\ct{$\pm~$}
\begin{table*}[t]
\centering
\caption{Overall accuracy (mean \ct standard deviation) (\%) of different dictionary learning methods on different datasets. Numbers in parentheses are number of training samples per class.}
\label{tab:overall_results}
{\small \begin{tabular}{|l||c|c|c|c|c|c|}
\hline
        & \begin{tabular}[c]{@{}c@{}}Ext. \\ YaleB (30) \end{tabular} & AR (20)    & \begin{tabular}[c]{@{}c@{}}AR \\ gender (250) \end{tabular} & \begin{tabular}[c]{@{}c@{}}Oxford \\ Flower (60) \end{tabular} & \begin{tabular}[c]{@{}c@{}}Caltech 101 \\ (30)\end{tabular} & COIL100 (10) \\  \hline \hline
SRC \cite{Wright2009SRC}              & 97.96 \ct 0.22           & 97.33 \ct 0.39           & 92.57 \ct 0.00          & 75.79 \ct 0.23          & 72.15 \ct 0.36          & 81.45 $\pm$ 0.80\\ \hline
LC-KSVD1 \cite{Zhuolin2013LCKSVD}     & 97.09 \ct 0.52           & 97.78 \ct 0.36           & 88.42 \ct 1.02          & 91.47 \ct 1.04                  & 73.40  \ct 0.64                 & 81.37 \ct 0.31\\ \hline
LC-KSVD2 \cite{Zhuolin2013LCKSVD}     & 97.80 \ct 0.37           & 97.70 \ct 0.23           & 90.14 \ct 0.45          & \textbf{\textit{92.00 \ct 0.73}} & 73.60 \ct 0.53                  & 81.42 \ct 0.33\\ \hline
DLSI \cite{ramirez2010classification} & 96.50 \ct 0.85           & 96.67 \ct 1.02           & 93.86 \ct 0.27          & 85.29 \ct 1.12          & 70.67 \ct 0.73           & 80.67 \ct 0.46  \\ \hline
DLRD~\cite{ma2012sparse}              & 93.56 \ct 1.25     & 97.83 \ct 0.80           & 92.71 \ct 0.43           &  -                      &  -                      &- \\ \hline
FDDL \cite{yang2014sparse}            & 97.52 \ct 0.63           & 96.16 \ct 1.16           & 93.70 \ct 0.24          & 91.17 \ct 0.89          & 72.94 \ct 0.26          & 77.45 \ct 1.04 \\ \hline
$D^2L^2R^2$\cite{li2014learning}      & 96.70 \ct 0.57           & 95.33 \ct 1.03           & 93.71 \ct 0.87          & 83.23 \ct 1.34          & 75.26 \ct 0.72          & 76.27 \ct 0.98\\ \hline
COPAR \cite{kong2012dictionary}       & \textbf{\textit{98.19 \ct 0.21}}  & 98.50 \ct 0.53  & \textbf{\textit{95.14 \ct 0.52}} & 85.29 \ct 0.74                  & \textbf{\textit{76.05 \ct 0.72}} & 80.46 \ct 0.61\\ \hline
JDL~\cite{zhou2014jointly}            & 94.99 \ct 0.53         & 96.00 \ct 0.96         & {93.86 \ct 0.43}        & 80.29 \ct 0.26        & {75.90 \ct 0.70}        & 80.77 \ct 0.85 \\ \hline
JDL$^*$~\cite{zhou2014jointly}        & 97.73 \ct 0.66         & \textbf{\textit{98.80 \ct 0.34}}         & 92.83 \ct 0.12       & 80.29 \ct 0.26        & 73.47 \ct 0.67        & 80.30 \ct 1.10\\ \hline
SRRS~\cite{li2016learning}            & 97.75 \ct 0.58         &   96.70 \ct 1.26                       & 91.28 \ct 0.15          & 88.52 \ct 0.64                         &   65.22 + 0.34                      & \bf{85.04 \ct 0.45}\\ \hline
LRSDL                                 & {\bf 98.76 \ct 0.23} & {\bf 98.87 \ct 0.43} & \bf{95.42 \ct 0.48}     & \bf{92.58 \ct 0.62}     & \bf{76.70 \ct 0.42}     & \textbf{\textit{84.35 \ct 0.37}}\\ \hline
\end{tabular}}
\end{table*}
\begin{figure}[t]
\centering
  \includegraphics[width=.4\textwidth]{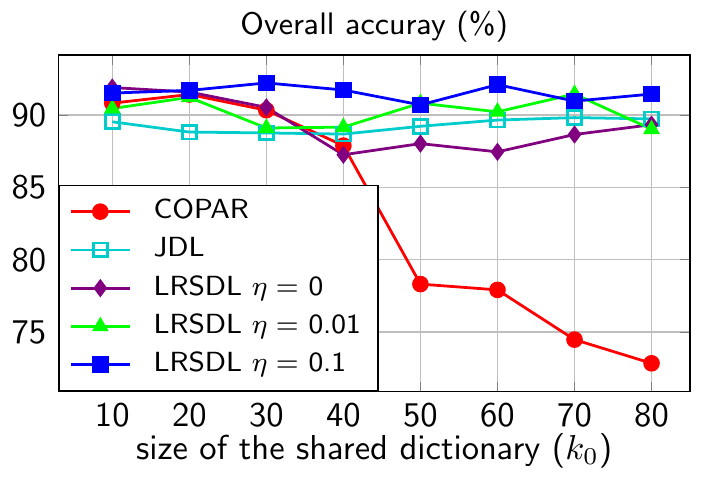}
 \caption{\small Dependence of overall accuracy on the shared dictionary.}
  \label{fig: compare_shared}
\end{figure}

\begin{figure}[t]
\centering
  \includegraphics[width=.4\textwidth]{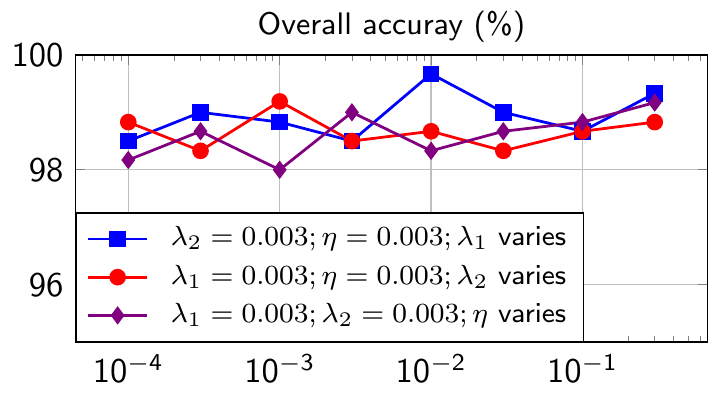}
 \caption{\small {Dependence of overall accuracy on parameters (the AR face dataset, $C = 100, n_c = 20, k_c = 15, k_0 = 10$).}}
  \label{fig:par_change}
\end{figure}
\subsection{Visualization of learned shared bases}
To demonstrate the behavior of dictionary learning methods on a dataset in the presence of shared features, we create a toy example in Figure \ref{fig: bases_simulated}. This is a classification problem with 4 classes whose basic class-specific elements and shared elements are visualized in Figure \ref{fig: bases_simulated}a). Each basis element has dimension $20$ pixel$\times 20$ pixel. From these elements, we generate 1000 samples per class by linearly combining class-specific elements and shared elements followed by noise added; 200 samples per class are used for training, 800 remaining images are used for testing. Samples of each class are shown in Figure \ref{fig: bases_simulated}b).

Figure \ref{fig: bases_simulated}c) show sample learned bases using DLSI \cite{ramirez2010classification} where shared features are still hidden in class-specific bases. In LC-KSVD bases (Figure \ref{fig: bases_simulated}d) and e)), shared features (the squared in the middle of a patch) are found
but they are classified as bases of class 1 or class 2, diminishing classification accuracy since most of test samples are classified as class 1 or 2. The same phenomenon happens in FDDL bases (Figure \ref{fig: bases_simulated}f)).
\par
The best classification results happen in three shared dictionary learnings (COPAR~\cite{kong2012dictionary} in Figure \ref{fig: bases_simulated}g), JDL~\cite{zhou2014jointly} in Figure \ref{fig: bases_simulated}h) and the proposed LRSDL in Figure \ref{fig: bases_simulated}i)) where the shared bases are extracted and gathered in the shared dictionary. However, in COPAR and JDL, shared features still appear in class-specific dictionaries and the shared dictionary also includes class-specific features. In LRSDL, class-specific elements and shared elements are nearly perfectly decomposed into appropriate sub dictionaries. The reason behind this phenomenon is the low-rank constraint on the shared dictionary of LRSDL. Thanks to this constraint, LRSDL produces perfect results on this simulated data.





\begin{figure*}[t]
\centering
  \includegraphics[width=\textwidth]{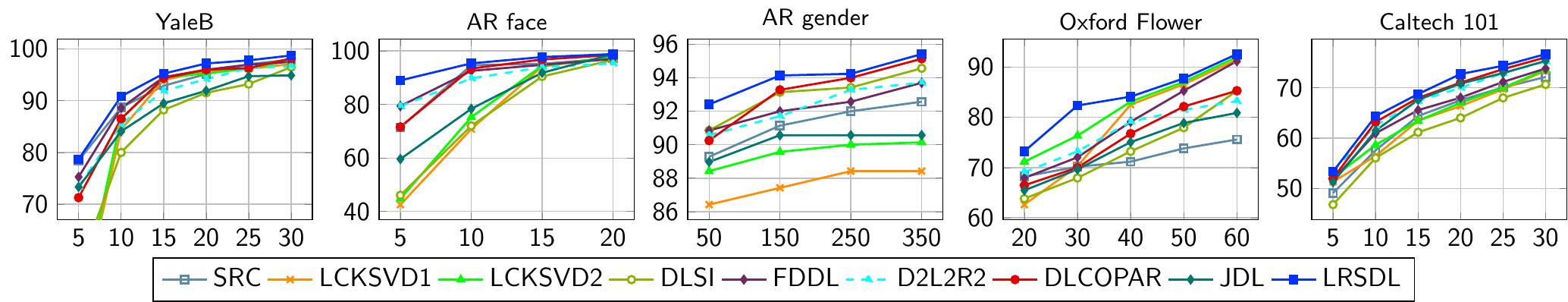}
  \vspace{-0.1in}
 \caption{\small Overall classification accuracy (\%) as a function of training set size per class.}
  \label{fig: compare_ntrains}
\end{figure*}
\vspace{-.2in}
\subsection{Effect of the shared dictionary sizes on overall accuracy} 
\vspace{-.1in}

We perform an experiment to study the effect of the shared dictionary size on the overall classification results of three shared dictionary methods: COPAR \cite{kong2012dictionary}, JDL~\cite{zhou2014jointly} and LRSDL in the AR gender dataset. In this experiment, 40 images of each class are used for training. The number of shared dictionary bases varies from 10 to 80. In LRSDL, because there is a regularization parameter $\eta$ which is attached to the low-rank term (see equation (\ref{eqn:lrsdl_cost_fn})), we further consider three values of $\eta$: $\eta = 0$, i.e.\ no low-rank constraint, $\eta = 0.01$ and $\eta=0.1$ for two different degrees of emphasis. Results are shown in Figure \ref{fig: compare_shared}.
\par

We observe that the performance of COPAR heavily depends on the choice of $k_0$ and its results worsen as the size of the shared dictionary increases. The reason is that when $k_0$ is large, COPAR tends to absorb class-specific features into the shared dictionary. This trend is {\em not} associated with LRSDL even when the low-rank constraint is ignored ($\eta = 0$), because LRSDL has another constraint ($\|\bX^0 - \bM^0\|_F^2$ small) which forces the coefficients corresponding to the shared dictionary to be similar. Additionally, when we increase $\eta$, the overall classification of LRSDL  also gets better. These observations confirm that our two proposed constraints on the shared dictionary are important, and the LRSDL exhibits robustness to parameter choices. {For JDL, we also observe that its performance is robust to the shared dictionary size, but the results are not as good as those of LRSDL.}


\vspace{-0.1in}
\subsection{Overall Classification Accuracy} 
\label{sub:overall_classification_acc}
Table \ref{tab:overall_results} shows overall classification results of various methods on all presented datasets {in terms of \textit{mean \ct standard deviation}}. It is evident that in most cases, three dictionary learning methods with shared features (COPAR \cite{kong2012dictionary}, JDL~\cite{zhou2014jointly} and our proposed LRSDL) outperform others with all five highest values presenting in our proposed LRSDL. {Note that JDL method represents the query sample class by class. We also extend this method by representing the query sample on the whole dictionary and use the residual for classification as in SRC. This extended version of JDL is called JDL*.}

\begin{table*}[]
\centering
{
\caption{Training and test time per sample (seconds) of different dictionary learning method on the Oxford Flower dataset ($n_c = 60, d = 10000, C = 17, K \approx 40\times 17$).} 
\label{tab:running_time}
\begin{tabular}{c|ccccccccccc}
\hline
      & SRC    & LCKSVD1 & LCKSVD2 & DLSI  & FDDL   & $D^2L^2R^2$  & COPAR & JDL    & SRRS & LRSDL \\ \hline
Train & 0      & 1.53e3    &  1.46e3   & 1.4e3 & 9.2e2  & $>$1 day      & 1.8e4 & 7.5e1  & 3.2e3     & 1.8e3\\ \hline
Test  & 3.2e-2 &  6.8e-3 &  6.1e-3 & 4e-3  & 6.4e-3 & 3.3e-2           & 5.5e-3& 3.6e-3 & 3.7e-3     & 2.4e-2\\ \hline
\end{tabular}}
\end{table*}

\vspace{-.2in}
\subsection{Performance vs. size of training set} 
\label{sub:performance_vs_size_of_training_set}
\vspace{-.1in}
Real-world classification tasks often have to contend with lack of availability of large training sets. To understand training dependence of the various techniques, we present a comparison of overall classification accuracy as a function of the training set size of the different methods. In Figure \ref{fig: compare_ntrains}, overall classification accuracies are reported for first five datasets\footnote{For the COIL-100 dataset, number of training images per class is already small (10), we do not include its results here.} corresponding to various scenarios. It is readily apparent that LRSDL exhibits the most graceful decline as training is reduced. In addition, LRSDL also shows high performance even with low training on AR datasets.
\vspace{-.2in}
{
\subsection{Performance of LRSDL with varied parameters}
\label{sub:par_change}
Figure \ref{fig:par_change} shows the performance of LRSDL on the AR face dataset with different values of $\lambda_1$, $\lambda_2$, and $\eta$ with other parameters fixed. We first set these three parameters to $0.003$ then vary each parameter from $10^{-4}$ to $0.3$ while two others are fixed. We observe that the performance is robust to different values with the accuracies being greater than 98\% in most cases. It also shows that LRSDL achieves the best performance when $\lambda_1 = 0.01, \lambda_2 = 0.003, \eta = 0.003$.
\subsection{Run time of different dictionary learning methods} 
\label{sub:running_time_of_different_dictionary_learning_methods}
Finally, we compare training and test time per sample of different dictionary learning methods on the Oxford Flower dataset. Note that, we use the efficient FDDL, DLSI, COPAR in this experiment. Results are shown in Table \ref{tab:running_time}. This result is consistent with the complexity analysis reported in Table \ref{tab:complexity_analysis} with training time of LRSDL being around half an hour, 10 times faster than COPAR~\cite{kong2012dictionary} and also better than other low-rank models, i.e. $D^2L^2R^2$~\cite{li2014learning}, and SRRS~\cite{li2016learning}. }
\section{Discussion and Conclusion} 
\label{sec:discussion_and_conclusion}
In this paper, our primary contribution is the development of a discriminative dictionary learning framework via the introduction of a shared dictionary {with two crucial constraints. First, the shared dictionary is constrained to be low-rank. Second, the sparse coefficients corresponding to the shared dictionary obey a similarity constraint.} In conjunction with discriminative model as proposed in \cite{Meng2011FDDL,yang2014sparse}, this leads to a more flexible model where shared features are excluded before doing classification. An important benefit of this model is the robustness of the framework to size ($k_0$) and the regularization parameter ($\eta$) of the shared dictionary. In comparison with state-of-the-art algorithms developed specifically for these tasks, our LRSDL approach offers better classification performance on average.
\par

In Section~\ref{sub:solving_the_opt} and \ref{sub:edlsi}, we discuss the efficient algorithms for FDDL \cite{yang2014sparse}, DLSI \cite{ramirez2010classification}, then flexibly apply them into more sophisticated models. Thereafter in Section~\ref{sec:complexity_analysis} and ~\ref{sub:valid_eff_algs}, we both theoretically and practically show that the proposed algorithms indeed significantly improve cost functions and run time speeds of different dictionary learning algorithms. The complexity analysis also shows that the proposed LRSDL requires less computation than competing models.
\par
As proposed, the LRSDL model learns a dictionary shared by every class. In some practical problems, a feature may belong to more than one but not all classes. Very recently, researchers have begun to address this issue \cite{yang2014latent,yoon2014hierarchical}.
In future work, we will investigate the design of hierarchical models for extracting common features among classes. 

\appendix
{\small \subsection{Proof of Lemma \ref{lem:fddl_updateD}}
\label{apd:proof_fddl_updateD}
 Let $\bw_c \in \{0, 1\}^{K}$ is a binary vector whose $j$-th element is one if and only if the $j$-th columns of $\bD$ belong to $\bD_c$, and $\bW_c = \diag(\bw_c)$. We observe that $\bD_c\bX^c_i = \bD\bW_c\bX_i$. We can rewrite $f_{\bY, \bX}(\bD)$ as:}
\small{ \begin{eqnarray*}
\nonumber
    &&\|\bY - \bD\bX\|_F^2 + \sum_{c=1}^C\big(\|\bY_c - \bD_c\bX^c_c\|_F^2 + \sum_{j\neq c} \|\bD_j\bX^j_c\|_F^2\big) \\
\nonumber
    &=& \|\bY - \bD\bX\|_F^2 + \sum_{c=1}^C\big(\|\bY_c - \bD\bW_c\bX_c\|_F^2 + \sum_{j\neq c} \|\bD\bW_j\bX_c\|_F^2\big), \\
    &=& \trace\left(\big( \bX\bX^T + \sum_{c=1}^C\sum_{j=1}^C\bW_j\bX_c\bX_c^T\bW_j^T\ \big)\bD^T\bD\right), \\
    && -2\trace\left(\big(\bY\bX^T + \sum_{c=1}^C \bY_c\bX_c^T\bW_c \big)\bD^T\right)+ \text{constant},\\
    &=& -2\trace(\bE\bD^T) + \trace(\Fb\bD^T\bD) + \text{constant}.
\end{eqnarray*}}
{ where we have defined:
\begin{eqnarray*}
    \bE &=& \bY\bX^T + \sum_{c=1}^C \bY_c\bX_c^T\bW_c,\\
    &=& \bY\bX^T + \bmt \bY_1(\bX_1^1)^T &\dots & \bY_C(\bX_C^C)^T\emt,  \\
    &=& \bY \left(\bX^T + \bmt
        (\bX_1^1)^T & \dots & \bzeros\\
        \bzeros  & \dots & \bzeros\\
        \dots    & \dots & \dots\\
        \bzeros  & \dots & (\bX_C^C)^T
        \emt\right) = \bY\M(\bX)^T,\\
    \Fb &=& \bX\bX^T + \sum_{c=1}^C\sum_{j=1}^C\bW_j\bX_c\bX_c^T\bW_j^T, \\
    &=& \bX\bX^T + \sum_{j=1}^C \bW_j \left(\sum_{c=1}^C\bX_c\bX_c^T\right) \bW_j^T,\\
    &=& \bX\bX^T + \sum_{j=1}^C \bW_j\bX\bX^T\bW_j^T.
\end{eqnarray*}
{Let:
\begin{equation*}
\bX\bX^T = \bA =
\bmt
        \bA_{11} &  \dots & \bA_{1j} & \dots & \bA_{1C}\\
        \dots    &  \dots & \dots    & \dots & \dots \\
        \bA_{21} &  \dots & \bA_{jj} & \dots & \bA_{2C}\\
        \dots    &  \dots & \dots    & \dots & \dots \\
        \bA_{C1} &  \dots & \bA_{Cj} & \dots & \bA_{CC}
        \emt.
\end{equation*}
From definition of $\bW_j$, we observe that `left-multiplying' a matrix by $\bW_j$ forces that matrix to be zero everywhere except the $j$-th block row. Similarly, `right-multiplying' a matrix by $\bW_j^T = \bW_j$ will keep its $j$-th block column only. Combining these two observations, we can obtain the result:
\begin{equation*}
    \bW_j\bA\bW_j^T =
    \bmt
        \bzeros &  \dots & \bzeros & \dots & \bzeros\\
        \dots    &  \dots & \dots    & \dots & \dots \\
        \bzeros &  \dots & \bA_{jj} & \dots & \bzeros\\
        \dots    &  \dots & \dots    & \dots & \dots \\
        \bzeros &  \dots & \bzeros & \dots & \bzeros
        \emt.
\end{equation*}
Then:
\begin{eqnarray*}
   \Fb &=&  \bX\bX^T + \sum_{j=1}^C \bW_j\bX\bX^T\bW_j^T, \\
   &=& \bA +  \bmt
        \bA_{11} & \dots & \bzeros\\
        \bzeros  & \dots & \bzeros\\
        \dots    & \dots & \dots\\
        \bzeros  & \dots & \bA_{CC}
        \emt = \M(\bA) = \M(\bX\bX^T).
\end{eqnarray*}
}
Lemma \ref{lem:fddl_updateD} has been proved. \hfill $\square$
\subsection{Proof of Lemma \ref{lem:fddl_updateX}}
\label{apd:proof_fddl_updateX}
We need to prove two parts:\\
\textit{For the gradient of} $f$, first we rewrite:
\begin{eqnarray}
\hspace{-0.1in}
\nonumber
  &&f(\bY, \bD, \bX) = \sum_{c=1}^C r(\bY_c, \bD, \bX_c) =\\
\nonumber
  && \norm{\underbrace{\bmt
    \bY_1 & \bY_2  & \dots & \bY_C \\
    \bY_1 & \mathbf{0} & \dots & \mathbf{0} \\
    \mathbf{0} & \bY_2 & \dots & \mathbf{0} \\
    \dots & \dots & \dots & \dots \\
    \mathbf{0} & \mathbf{0} & \dots & \bY_C
    \emt}_{\widehat{\bY}}
   - \underbrace{\bmt
     \bD_1 & \bD_2  & \dots & \bD_C \\
     \bD_1 & \mathbf{0} & \dots & \mathbf{0} \\
     \mathbf{0} & \bD_2 & \dots & \mathbf{0} \\
     \dots & \dots & \dots & \dots \\
     \mathbf{0} & \mathbf{0} & \dots & \bD_C
     \emt}_{\widehat{\bD}}
  \bX
  }_F^2\\
\nonumber
  &&= \| \widehat{{\bY}}  - \widehat{\bD} \bX \|_F^2.
\end{eqnarray}
Then we obtain:}
{ \begin{eqnarray*}
     \frac{\partial \frac{1}{2}f_{\bY, \bD}(\bX)}{\partial \bX}
     = \widehat{\bD}^T\widehat{\bD} - \widehat{\bD}^T\widehat{\bY} = \M(\bD^T\bD) \bX - \M(\bD^T\bY).
\end{eqnarray*}
\textit{For the gradient of} $g$, let $\bE_{p}^q$ be the all-one matrix in $\R^{p\times q}$. It is easy to verify that:}
{ \begin{eqnarray*}
    (\bE_p^q)^T = \bE_q^p,
    & &
    \bM_c = \bm_c\bE_{1}^{n_c} = \frac{1}{n_c}\bX_c\bE_{n_c}^{n_c},
    \\
    \bE_p^q\bE_q^r = q\bE_p^r,
    &&
    (\bI-\frac{1}{p}\bE_p^p)(\bI - \frac{1}{p}\bE_p^p)^T = (\bI - \frac{1}{p}\bE_p^p).
\end{eqnarray*}}
 We have:\\
	\begin{equation*}
	         \bX_c - \bM_c = \bX_c - \frac{1}{n_c}\bX_c\bE_{n_c}^{n_c} = \bX_c(\bI - \frac{1}{n_c}\bE_{n_c}^{n_c}),
	\end{equation*}
    \begin{eqnarray*}
          \imply \frac{\partial}{\partial \bX_c}\frac{1}{2}\|\bX_c - \bM_c\|_F^2 &=&   \bX_c (\bI - \frac{1}{n_c}\bE_{n_c}^{n_c})(\bI - \frac{1}{n_c}\bE_{n_c}^{n_c})^T, \\
          																		&=& \bX_c (\bI - \frac{1}{n_c}\bE_{n_c}^{n_c}) = \bX_c - \bM_c.
      \end{eqnarray*}
Therefore we obtain:

    \begin{align}
    \nonumber
        \frac{\partial \frac{1}{2}\sum_{c=1}^C \|\bX_c - \bM_c\|_F^2}{\partial \bX} &= [ \bX_1, \dots, \bX_C] - \underbrace{[ \bM_1, \dots, \bM_C]}_{\widehat{\bM}} \\
        \label{eqn:dif_g_1}
        &= \bX - \widehat{\bM}.
    \end{align}
    For $\bM_c - \bM$, first we write it in two ways:
{ \small  \begin{align}
        \nonumber
        &\bM_c - \bM = \frac{1}{n_c} \bX_c \bE_{n_c}^{n_c} - \frac{1}{N}\bX\bE_{N}^{n_c}
         =  \frac{1}{n_c} \bX_c \bE_{n_c}^{n_c} - \frac{1}{N}\sum_{j=1}^C\bX_j\bE_{n_j}^{n_c}, \\
        \label{eqn:mcm_xc}
        & = \frac{N - n_c}{Nn_c}\bX_c \bE_{n_c}^{n_c} - \frac{1}{N}\sum_{j \neq c}\bX_j\bE_{n_j}^{n_c},\\
        \label{eqn:mcm_xj}
        & = \frac{1}{n_c} \bX_c \bE_{n_c}^{n_c} - \frac{1}{N}\bX_l\bE_{n_l}^{n_c} - \frac{1}{N}\sum_{j\neq l}\bX_j\bE_{n_j}^{n_c}  ~~ (l\neq c).
    \end{align}}
    Then we infer:
    { \begin{align*}
    \displaystyle
        \nonumber
            (\ref{eqn:mcm_xc}) \imply  \frac{\partial}{\partial \bX_c} \frac{1}{2} \|\bM_c - \bM\|_F^2 &= \left(\frac{1}{n_c} - \frac{1}{N}\right)(\bM_c - \bM)\bE_{n_c}^{n_c}, \\
                & = (\bM_c - \bM) + \frac{1}{N} (\bM - \bM_c)\bE_{n_c}^{n_c}.\\
           (\ref{eqn:mcm_xj}) \imply  \frac{\partial}{\partial \bX_l} \frac{1}{2} \|\bM_c - \bM\|_F^2 &= {\frac{1}{N}(\bM - \bM_c)\bE_{n_c}^{n_l}} (l \neq c).
       \end{align*}}

        {\small $\displaystyle\imply\frac{\partial}{\partial \bX_l} \frac{1}{2}\sum_{c=1}^C \|\bM_c - \bM\|_F^2 = \bM_l - \bM + \frac{1}{N}\sum_{c=1}^C (\bM - \bM_c)\bE_{n_c}^{n_l}.$}\\
{    Now we prove that $\displaystyle\sum_{c=1}^C (\bM - \bM_c)\bE_{n_c}^{n_l} = \bzeros$. Indeed,
    \begin{align*}
    	&\sum_{c=1}^C (\bM - \bM_c)\bE_{n_c}^{n_l}= \sum_{c=1}^C(\bm\bE_{1}^{n_c} - \bm_c\bE_{1}^{n_c})\bE_{n_c}^{n_l}, \\
    	&= \sum_{c=1}^C(\bm - \bm_c)\bE_{1}^{n_c}\bE_{n_c}^{n_l}
    	=\sum_{c=1}^Cn_c(\bm - \bm_c)\bE_{1}^{n_l,} \\
    	&= \big(\sum_{c=1}^C n_c \bm - \sum_{c=1}^C n_c\bm_c\big)\bE_{1}^{n_l} =\bzeros~\left(\text{since~} \bm = \frac{\sum_{c=1}^C n_c \bm_c}{\sum_{c=1}^C n_c}\right).
    \end{align*}
    Then we have:}
 { \small   \begin{align}
        \label{eqn:dif_g_2}
        \frac{\partial}{\partial \bX}\frac{1}{2}\sum_{c=1}^C \|\bM_c - \bM\|_F^2 = [\bM_1, \dots, \bM_C] - \bM = \widehat{\bM} - \bM.
    \end{align}
    Combining (\ref{eqn:dif_g_1}), (\ref{eqn:dif_g_2}) and $\frac{\partial \frac{1}{2}\|\bX\|_F^2}{\partial \bX} = \bX$ , we have:}
{    \begin{equation*}
        \frac{\partial \frac{1}{2}g(\bX)}{\partial \bX} = 2\bX + \bM - 2\widehat{\bM.}
    \end{equation*}
    Lemma \ref{lem:fddl_updateX} has been proved. \hfill $\square$
\subsection{Proof of Lemma \ref{lem:lrsdld0x0}}
\vspace{-0.1in}
\label{apd:proof_lrsdl_updateD0X0}
When $\bY, \bD, \bX$ are fixed, we have:
\begin{align}
\label{eqn:sddl_solve_X0_ori}
    \nonumber
    &J_{\bY, \bD, \bX}(\bD_0, \bX^0) = \frac{1}{2}\|\bY - \bDc\bX^0 - \bD\bX\|_F^2 + \eta\|\bD_0\|_*+ \\
        &\sum_{c=1}^{C}\frac{1}{2}\|\bY_c - \bDc\bX_c^0 - \bD_c\bX_c^c\|_F^2
        +\lambda_1\|\bX^0\|_1  + \text{constant.}
\end{align}}

Let $\tilde{\bY} = \bY - \bD\bX$, $\hat{\bY}_c = \bY_c - \bD_c \bX_c^c$ and $\hat{\bY} =
    \bmt \hat{\bY}_1 & \hat{\bY}_2 & \dots & \hat{\bY}_C \emt $, we can rewrite (\ref{eqn:sddl_solve_X0_ori}) as:

{\begin{eqnarray*}
\nonumber
 J_{\bY, \bD, \bX}(\bD_0, \bX^0) =\frac{1}{2}\|\tilde{\bY} - \bDc\bXc\|_F^2 + \frac{1}{2}\|\hat{\bY} - \bDc\bXc \|_F^2 + \\
  +\lambda_1 \|\bXc\|_1 + \eta\|\bD_0\|_* + \text{constant}_1,\\
    = \norm{\frac{\tilde{\bY} + \hat{\bY}}{2} - \bDc\bXc}_F^2 + \lambda_1 \|\bXc\|_1 + \eta\|\bD_0\|_* + \text{constant}_2.
\end{eqnarray*}
We observe that:
 \begin{eqnarray*}
    \tilde{\bY} + \hat{\bY} &=& 2\bY - \bD\bX - \bmt \bD_1\bX_1^1 &\dots & \bD_C\bX_C^C\emt \\
    &=& 2\bY - \bD\M(\bX).
\end{eqnarray*}
Now, by letting
$\displaystyle\bV = \frac{\tilde{\bY} + \hat{\bY}}{2} $, Lemma \ref{lem:lrsdld0x0} has been proved. $~~~$} \hfill $\square$
\bibliographystyle{myIEEEtran}
\bibliography{LRSDL_accepted}

\begin{IEEEbiography}[{\includegraphics[width=1in,height=1.25in,clip,keepaspectratio]{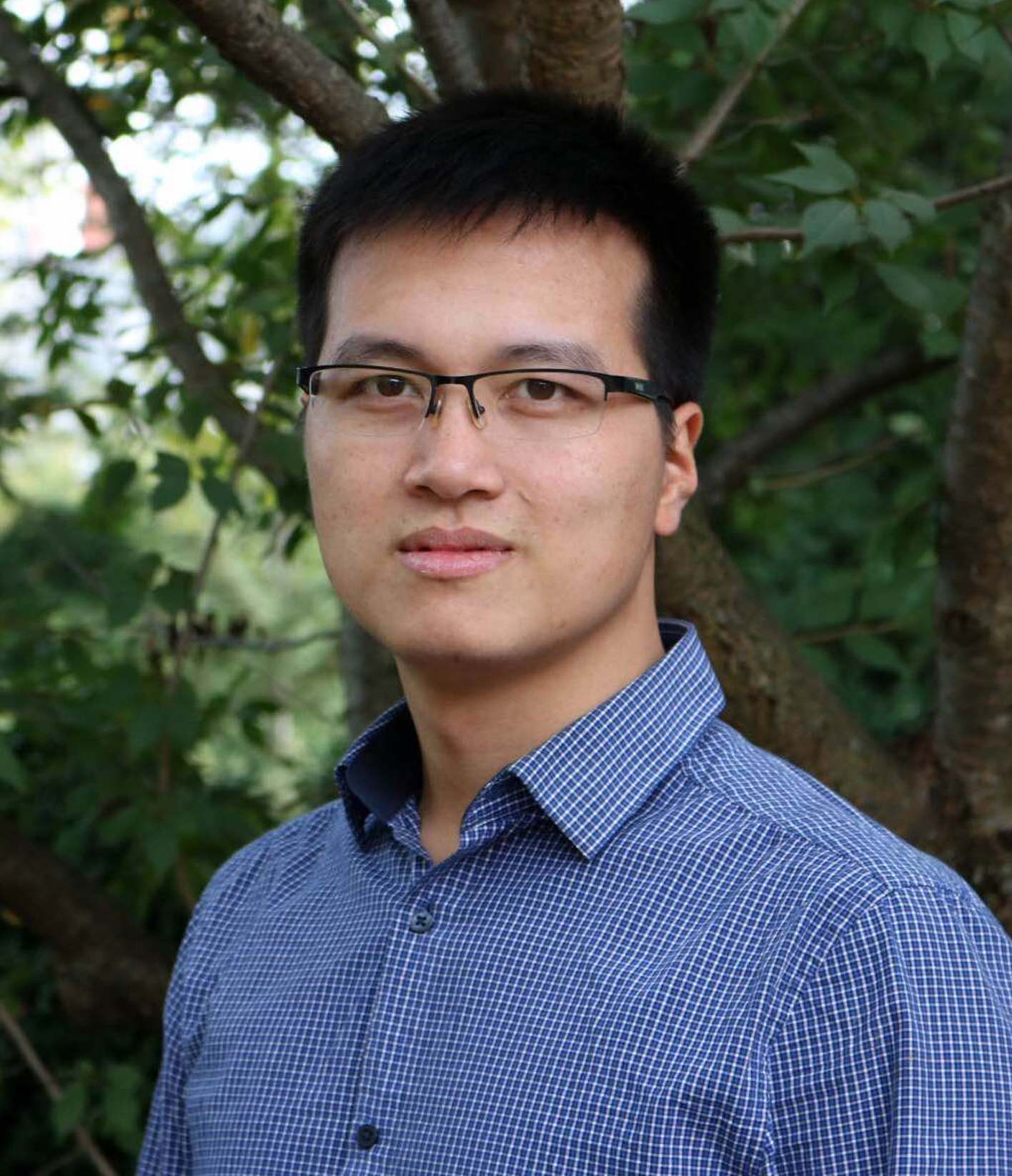}}]{Tiep Huu Vu} received the B.S. degree in Electrical Engineering from Hanoi University of Science and Technology, Vietnam, in 2012. He is currently pursuing the Ph.D. degree with the information Processing and Algorithm Laboratory (iPAL), The Pennsylvania State University. His research interests are broadly in the areas of statistical learning for signal and image analysis, computer vision and pattern recognition for image classification, recovery and retrieval. 

\end{IEEEbiography}

\begin{IEEEbiography}[{\includegraphics[width=1in,height=1.25in,clip,keepaspectratio]{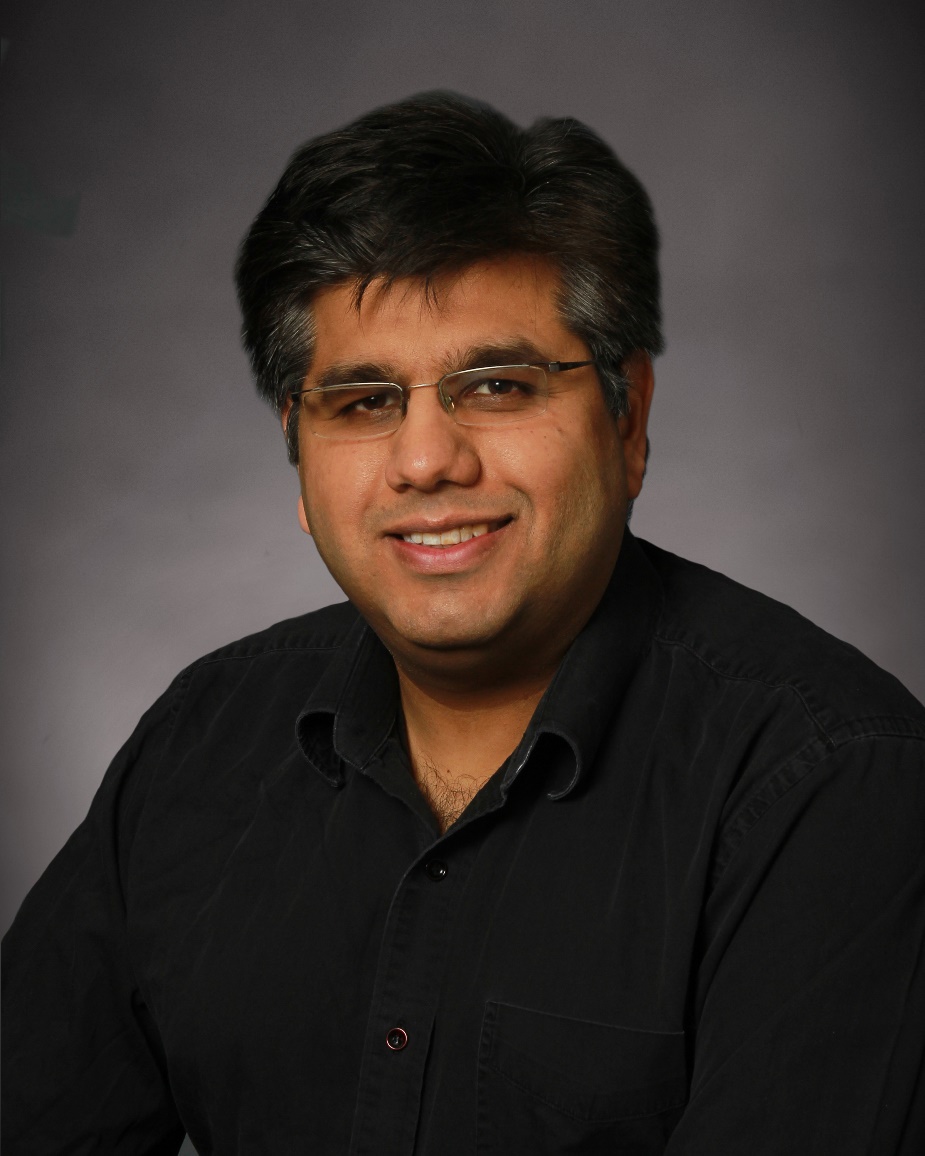}}]{Vishal Monga}
(SM  2011) is a tenured Associate Professor in the School of Electrical Engineering and Computer Science at the main campus of Pennsylvania State University in University Park, PA. He was with Xerox Research from 2005-2009. His undergraduate work was completed at the Indian Institute of Technology (IIT), Guwahati and his doctoral degree in Electrical Engineering was obtained from the University of Texas, Austin in Aug 2005.
Dr. Monga’s research interests are broadly in signal and image processing.  His research group at Penn State focuses on convex optimization approaches to image classification, robust signal (time-series) hashing, radar signal processing and computational imaging.
He currently serves as an Associate Editor for the IEEE Transactions on Image Processing, IEEE Signal Processing Letters, and the IEEE Transactions on Circuits and Systems for Video Technology. Prof. Monga is a recipient of the US National Science Foundation (NSF) CAREER award, a Monkowski Early Career award from the college of engineering at Penn State and the Joel and Ruth Spira Teaching Excellence Award.

\end{IEEEbiography}

\end{document}